\pdfoutput=1
\documentclass[11pt]{article}
 
\usepackage[]{ACL2023} 

\usepackage{times}
\usepackage{latexsym}
\usepackage{amsmath}
\usepackage[T1]{fontenc}
\usepackage[utf8]{inputenc}

\usepackage{microtype}

\usepackage{inconsolata}

\usepackage{pdfpages}

\usepackage{csquotes}
\usepackage{multirow}

\usepackage{pifont}
\usepackage{xcolor}
\newcommand{\cmark}{\textcolor{green!80!black}{\ding{51}}}
\newcommand{\xmark}{\textcolor{red}{\ding{55}}}

\usepackage{makecell}

\usepackage{algorithm}
\usepackage{algpseudocode}

\usepackage{ifthen}
 \newboolean{showcomments}
 \setboolean{showcomments}{true} % Set to false to hide all comments
\newcommand{\alicja}[1]{\ifthenelse{\boolean{showcomments}}{\textcolor{cyan}{[#1 —alicja]}}{}}
\newcommand{\raj}[1]{\ifthenelse{\boolean{showcomments}}{\textcolor{green}{[#1 —raj]}}{}}
\newcommand{\rajbrown}[1]{\ifthenelse{\boolean{showcomments}}{\textcolor{brown}{[#1 —raj]}}{}}
\newcommand{\diyi}[1]{\ifthenelse{\boolean{showcomments}}{\textcolor{blue}{[#1 —diyi]}}{}}
\newcommand{\ryan}[1]{\ifthenelse{\boolean{showcomments}}{\textcolor{orange}{[#1 —ryan]}}{}}
\newcommand{\bob}[1]{\ifthenelse{\boolean{showcomments}}{\textcolor{red}{[#1 —bob]}}{}}

\definecolor{myColour}{HTML}{003366}
\makeatletter
\def\thanks#1{\protected@xdef\@thanks{\@thanks
        \protect\footnotetext{#1}}}
\makeatother
\title{Multi-Level Feedback Generation with Large Language Models for Empowering Novice Peer Counselors}

\newcommand{\gtlogo}{\raisebox{3.4pt}{\includegraphics[scale=0.04]{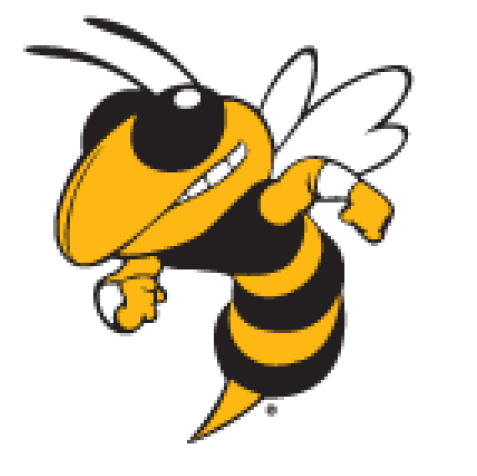}}}
\newcommand{\stanfordlogo}{\raisebox{3.4pt}{\includegraphics[scale=0.04]{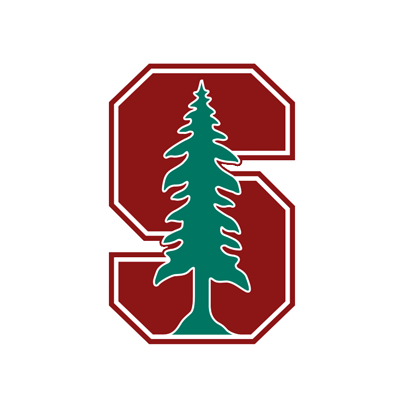}}}
\newcommand{\cmulogo}{\raisebox{3.4pt}{\includegraphics[scale=0.035]{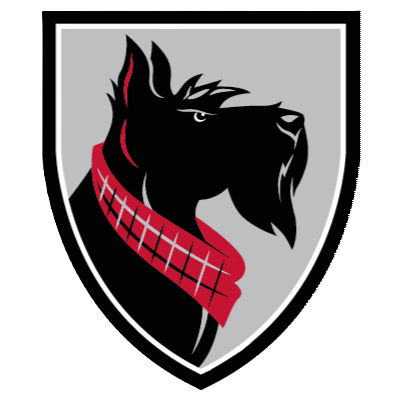}}}

\author{\bf \hypersetup{linkcolor=black} Alicja Chaszczewicz \stanfordlogo,
        Raj Sanjay Shah* \gtlogo,
        Ryan Louie* \stanfordlogo\\
        \bf \hypersetup{linkcolor=black} Bruce A Arnow \stanfordlogo,
        Robert Kraut \cmulogo, 
        Diyi Yang \stanfordlogo \\
        Stanford University \stanfordlogo, 
        Georgia Institute of Technology \gtlogo,
        Carnegie Mellon University \cmulogo \\
        \thanks{Email IDs of the authors: \{alicjach, rylouie, arnow, diyiy\}@stanford.edu, rajsanjayshah@gatech.edu, robert.kraut@cmu.edu}\thanks{* These authors contributed equally to this work}
        }

\definecolor{ReflectionsColor}{RGB}{204, 229, 255}
\definecolor{QuestionsColor}{RGB}{255, 204, 204}
\definecolor{SuggestionsColor}{RGB}{204, 255, 204}
\definecolor{ValidationColor}{RGB}{255, 255, 204}
\definecolor{SelfDisclosureColor}{RGB}{204, 204, 255}
\definecolor{EmpathyColor}{RGB}{255, 229, 204}
\definecolor{ProfessionalismColor}{RGB}{229, 204, 255}
\definecolor{StructureColor}{RGB}{204, 255, 229}

\begin{document}
\maketitle
\begin{abstract}
% \diyi{please do not accept my edits, and keep them on for now.}

Realistic practice and tailored feedback are key processes for training peer counselors with clinical skills. However, existing mechanisms of providing feedback largely rely on human supervision. Peer counselors often lack mechanisms to receive detailed feedback from experienced mentors, making it difficult for them to support the large number of people with mental health issues who use peer counseling.
Our work aims to leverage large language models to provide contextualized and multi-level feedback to empower peer counselors, especially novices, at scale.
To achieve this, we co-design with a group of senior psychotherapy supervisors to develop a multi-level feedback taxonomy, and then construct a publicly available dataset with comprehensive feedback annotations of 400 emotional support conversations. We further design a self-improvement method on top of large language models to enhance the automatic generation of feedback. 
Via qualitative and quantitative evaluation with domain experts,
we demonstrate that our method minimizes the risk of potentially harmful and low-quality feedback generation which is desirable in such high-stakes scenarios.

\end{abstract}

\section{Introduction}
 
Realistic practice and tailored feedback are key processes for training peer counselors with clinical skills. Providing feedback could significantly enhance peer counselor skills, thereby improving support quality and benefiting many seeking help online~\cite{ali2015online}. However, it is often time-consuming and costly for counseling supervisors to provide detailed feedback \cite{atkins2014scaling} to beginner peer counselors.
Without appropriate guidance, peer counselors might develop biased or even inappropriate helping skills without

\begin{figure}[H]
\centering
\includegraphics[width=0.48\textwidth]{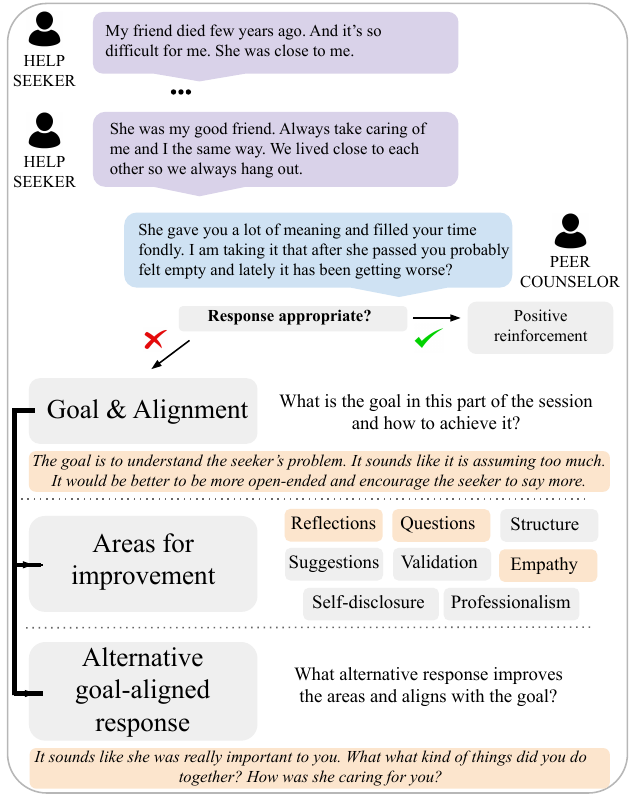}
\caption{Example conversation excerpt taken from the ESConv dataset \cite{liu-etal-2021-towards} annotated using our feedback taxonomy. Feedback components (\textit{appropriateness, goal definition and alignment, areas for improvement, alternative goal-aligned response}) are demonstrated on one utterance of peer counselor's response (in blue). Optionally, one can also provide \textit{positive reinforcement} by highlighting areas in categories peer counselors excelled at.}
\label{fig:feedback_framework}
\end{figure}

\noindent being aware of it, based on their own experiences. What can we do to provide detailed feedback to a large number of novice peer counselors at scale? In this work, we explore whether large language models (LLMs) can be used to provide contextualized feedback to empower peer counselors in training.

\if 0

With the goal of mimicking human supervision experience, we interview senior psychotherapy supervisors of counselors in training to understand the content and delivery of feedback to novice counselors. 
In the first step of the co-design process, we conduct a contextual inquiry 
\citep{Karen2017-fp} on supervisors engaging in a representative task of providing feedback on a transcript of an emotional support conversation \cite{liu-etal-2021-towards} as if they were communicating the feedback to a novice. We then develop our feedback framework by modeling the common patterns at different granularity observed in interviews and including important feedback dimensions highlighted in textbooks and training for foundational active listening skills \cite{hill2009helping, 7cups2023}. Aiming to make a step toward how feedback is delivered in the professional supervision of peer counselors, we explore whether large language models (LLMs) can be used to provide contextualized feedback to empower novice counselors.
\fi

Numerous recent studies have explored the feasibility of applying computational techniques to differentiate between low and high-quality counseling automatically \cite{perez-rosas-etal-2019-makes, RN9275, sharma-etal-2020-computational, flemotomos2021automated, min-etal-2022-pair,shen-etal-2022-knowledge, fi15030110, fang2023makes, sharma2023human, hsu2023helping, chiu2024computational}. In doing so, prior work mostly provides numeric feedback to counselors about how well a particular skill is used.
Some recent studies provide utterance-level suggestions of responses to use according to appropriate helping skills \cite{hsu2023helping}, or alternatives for more empathetic responses \cite{sharma2023human}. 
Yet, little attention is given to developing automatic feedback that closely mirrors how clinical supervisors provide feedback to novice counselors.

To this end, we co-designed a feedback framework with senior psychotherapy supervisors to reflect the content and delivery of feedback they give to novice counselors. 
Concretely, we conducted a contextual inquiry 
\citep{Karen2017-fp} with supervisors engaging in a representative task of providing feedback on a transcript of an emotional support conversation \cite{liu-etal-2021-towards} as if they were communicating the feedback to a novice counselor. We then developed a multi-level feedback framework by modeling the common patterns at different granularity observed in interviews
and important feedback dimensions highlighted in textbooks and training for foundational active listening skills \cite{hill2009helping, 7cups2023}.
With this multi-level feedback framework presented in Figure \ref{fig:feedback_framework}, we introduce a publicly available dataset of conversations enriched with comprehensive feedback annotations, building upon an existing public emotional support conversations dataset ESConv \cite{liu-etal-2021-towards}. Specifically, we leverage a model-in-the-loop annotation paradigm where GPT-4 and counseling domain experts work together to produce the annotations for 400 conversations.

To enable transparent model development, especially for a high-stakes domain like counseling, we fine-tuned the open-source Llama-2 model to generate multi-level feedback. We further introduce a simple but effective self-improvement method to forecast how specific feedback might improve subsequent interaction and use this forecast information to supervise feedback generation. Unlike general natural language generation tasks, we aim at optimizing feedback generation for worst-case performance since failures (e.g., generating poor advice) matter more in this high-stakes scenario. Using both quantitative evaluation and qualitative evaluation with domain experts, we demonstrate that our approach generates high-quality feedback and significantly boosts the worst-case performance on multi-level feedback generation compared to baselines. 
In summary, this paper makes the following contributions:

\begin{itemize}\itemsep0em 
    \item We propose a novel and comprehensive multi-level feedback framework for training peer counseling skills co-designed with senior psychotherapy supervisors. 
    \item We constructed and make publicly available \emph{FeedbackESConv}\footnote{We will release our code at https://github.com/SALT-NLP/counseling-feedback}, a dataset of 400 emotional support conversations with multi-level feedback annotated by domain experts and GPT-4.
    \item We enhanced a fine-tuned LLM for multi-level feedback using a simple but effective self-improvement method 
    to forecast how specific feedback might improve subsequent interaction and further use such signal to supervise the feedback generation. 
    \item We conducted extensive evaluations with domain experts to demonstrate the effectiveness of our method and find that, compared to baselines, it significantly boosts the worst-case performance on multi-level feedback generation. 
\end{itemize}

\begin{table*}[t]
  \centering
  \resizebox{\textwidth}{!}{
\begin{tabular}{|c|c|c|c|c|}  
    \cline{2-5}
    \multicolumn{1}{c|}{} & \textbf{\small{\makecell{Numerical scoring \\ of response quality}}} & \textbf{\small{\makecell{Suggestion of response \\ or alternate response}}} & \textbf{\small{\makecell{Response evaluation\\ across multiple peer \\ counseling skills categories}}} &   \textbf{\small{\makecell{Goal - oriented \\ natural language \\ explanations}}}   \\
    \hline
     \small{\citet{perez-rosas-etal-2019-makes}} & \cmark & \xmark & \cmark &  \xmark\\
     \hline
    \small{\citet{tanana2019development, RN9275}}& \cmark & \xmark & \cmark &  \xmark\\
    \hline
    \small{\citet{sharma-etal-2020-computational}} & \cmark & \xmark & \xmark & \xmark\\ 
     \hline
   \small{\citet{flemotomos2021automated}} & \cmark & \xmark & \cmark &  \xmark\\
    \hline
    \small{\citet{min-etal-2022-pair}} & \cmark & \xmark & \xmark & \xmark\\
     \hline
    \small{ \citet{shen-etal-2022-knowledge}} &  \xmark & \cmark & \xmark & \xmark\\
    \hline
    \small{\citet{sharma2023human}} &  \xmark & \cmark & \xmark & \xmark\\
    \hline
    \small{\citet{hsu2023helping}} &   \xmark & \cmark & \cmark & \xmark\\
    \hline
    \small{\citet{chiu2024computational}*} & \cmark & (\cmark) & \cmark & \xmark\\
    \hline
    \small{Our work} & \cmark & \cmark & \cmark & \cmark\\
    \hline
\end{tabular}
}

\caption{Categorization of previously proposed approaches aimed at evaluating or enhancing the quality of emotional support conversations. "Numerical scoring of response quality" indicates whether a study applied a binary or continuous scale for quality assessment. "Response evaluation across multiple peer counseling skills categories" indicates whether the feedback mechanism incorporated a multidimensional structure (more than two dimensions). "Suggestion of response" examines if the approach includes generating potential peer counselor answers. "Goal-oriented natural language explanation" indicates whether the system offers natural language conversation goals and explains how errors it identified can be aligned to these goals. *\citet{chiu2024computational} is concurrent work focusing on evaluating the quality of LLM-based therapy simulations.}
\label{tab:feedback}
\end{table*}

\section{Related Work}
\label{sec:related_work}

\subsection{Automated Feedback for Peer Counseling}
\label{sec:related_work_feedback}
There have been different approaches to build automated methods that help peer counselors evaluate and improve their skills, ranging from scoring-based methods (e.g., measures of empathy; the use of counseling-specific dialogue acts~\cite{sharma-etal-2020-computational,min-etal-2022-pair, perez-rosas-etal-2019-makes, tanana2019development, flemotomos2021automated, chiu2024computational} to automatically generated-suggestions for alternative responses \cite{shen-etal-2022-knowledge,sharma2023human, hsu2023helping}.

Rather than taking a technical perspective focusing on the feedback systems that \emph{can} be built with scoring or response generation methods, we posit that one can design better-automated feedback methods for peer counseling training by understanding and mirroring the existing ways supervisors deliver feedback to novices. Thus, in this work, we take a collaborative design approach with senior psychotherapy supervisors who are experienced in giving tailored feedback to novice counselors. We translate their input into the peer counseling domain and use it to inform the construction of our multi-level feedback taxonomy.
 
Our co-design reveals that post-session feedback for peer counseling \emph{encompasses and extends beyond} scoring and suggestions for improving the quality of individual responses. Most differently, it emphasizes that each response should be based on the counseling goals it should serve at the specific point in the session. Incorporating contextualized \emph{goals} into the feedback structure provides a purpose-led orientation compared to previous approaches (see Table \ref{tab:feedback}). Natural language goal descriptions are especially valuable since providing explanations is more beneficial for learning than simply giving the correct answer~\citep{butler2013explanation}.

\if 0

We focus on generating feedback with LLMs to help peer counselors master counseling skills; similar avenues have been explored in other contexts, for example, in math tutoring \cite{wang2023step} or in writing of research papers \cite{liang2023can}. 
\fi

\if 0
The details of this framework, co-designed with senior counseling supervisors, are presented in Section \ref{sec:feedback_framework}.
\fi

\subsection{Generation Capabilities of LLMs}
Past work explored the capabilities of LLMs in generating natural-language feedback across various domains.  ~\citet{wang2023step} explores the use of LLMs like GPT-4 and GPT3.5 in math tutoring to deliver high-quality feedback to remediate student mistakes. \citet{liang2023can} employ GPT-4 for generating comprehensive reviews for research papers. These varied applications demonstrate the adaptability and potential of LLMs to generate feedback across educational and professional settings. Unlike past work that builds feedback systems directly on top of GPT-4, we seek to enable the transparent development of open-source feedback models for the domain of peer counseling. Thus, we first develop an annotated dataset of feedback which is co-annotated by domain experts and GPT-4 using our multi-level feedback taxonomy, and then fine-tune the open-source Llama2-13B model using this feedback dataset.

The effectiveness of LLM feedback, and of LLM generated outputs more broadly, can be undermined by undesired and inconsistent behaviors, including hallucination, unfaithful reasoning, and toxic content. A promising approach to rectify these flaws is using self-correction or self-improvement techniques, in which a source of automated feedback, either produced by the LLM itself or some external system, can prompt or guide the LLM to fix problems in its output~\cite{pan2023automatically}.  Self-correction methods can be categorized into training-time, generation-time, and post-hoc corrections. 
Our self-improvement method is most related to training-time self-corrections. For example, \citet{huang-etal-2023-large} used self-consistency~\cite{wang2023selfconsistency} and chain of thought (CoT) prompting to select best generations for further supervised fine-tuning (SFT) on reasoning tasks. \citet{selfee2023} fine-tuned LLama models with self-feedback and revision data generated by ChatGPT to enable the model to self-revise its outputs. Concurrent to our work,~\citet{yuan2024self} uses iterative LLM-as-a-Judge~\cite{zheng2023judging} prompting to obtain self-rewards and perform direct preference optimization~\cite{rafailov2023direct} to perform model alignment to the preferences from this self-reward.

In our work, undesirable and inconsistent LLM feedback generation may include poor goal identification or utterance-level rewrites that are inconsistent with the conversation goals. To mitigate this, we developed a training-time self-improvement method that relies on the fine-tuned LLM itself to provide automated scoring feedback on candidate outputs; this allows it to select preferred generations upon which the feedback model can be further preference-tuned.  

\if 0
Our work explores an LLM's ability to generate multi-level feedback, which requires reasoning about contextualized goals and producing utterance-level rewrites aligned to those goals. The efficacy of LLM feedback generation, however, can be undermined by undesired and inconsistent behaviors, such as poor goal identification or utterance level rewrites that are inconsistent with these goals.
To mitigate these problems, we developed a self-improvement method
inspired by the successes of similar self-correction methods that rely on the LLM itself to provide automated feedback~\citep{pan2023automatically}. Self-correction methods can be categorized into training-time, generation-time, and post-hoc corrections. Our self-improvement method is most related to training-time self-corrections. For example, \citet{huang-etal-2023-large} used self-consistency~\cite{wang2023selfconsistency} and chain of thought (CoT) prompting to select best generations for further supervised fine-tuning (SFT) on reasoning tasks. \citet{selfee2023} fine-tuned LLama models with self-feedback and revision data generated by ChatGPT to enable the model to self-revise its outputs. Concurrent to our work,~\citet{yuan2024self} uses iterative LLM-as-a-Judge~\cite{zheng2023judging} prompting to obtain self-rewards and perform DPO~\cite{rafailov2023direct} on preferences from this self-reward. 

\fi

\if 0
\ryan{currently commenting this out, since the list of self-correction methods here felt like a laundry list and was not organized as well}
\fi

\section{Feedback Framework}
\label{sec:feedback_framework}
\if 0
\fi

Given the crucial role of human supervision and tailored contextual feedback in the peer counselors training process \cite{borders2005new, bernard1998fundamentals, gonsalvez2010clinical, ronnestad2013developing}, we collaborated with senior psychotherapy supervisors (each with over 20 years of experience) to develop an automated feedback system that is aligned with best peer counseling practices. Together, we co-designed a multi-level feedback framework for peer counselor training. 

Four one-hour co-design sessions with these senior supervisors revealed that initial training of novice therapists emphasizes foundational active listening skills and that these are generic skills common to all therapy approaches, including peer counseling \cite{watkins2014wiley, RN9183, RN9184, RN9115}. Details about the co-design process including research questions, key themes, and the outcomes are given in  Appendix \ref{sec:appendix_interview}.
Via our co-design, we found that the structure of the supervisors' feedback spans different levels: it often starts with positive reinforcement, followed by a line-by-line analysis of session transcripts; for any utterances needing improvement, supervisors clarified the session goals, identified categories of skills that could be improved, and voiced alternative responses that would achieve the goals.
    
\subsection{Multi-Level Feedback Taxonomy}
\label{sec:feedback_components}
Building upon our co-design sessions, we derive a multi-level feedback framework that reflects the components of senior psychotherapy supervisors' feedback and trains foundational listening skills that are relevant to peer counseling; see Figure~\ref{fig:feedback_framework}. This taxonomy has five key components: 
\begin{enumerate}\itemsep0em
    \item \textbf{Appropriateness} indicates whether a peer counselor's response in a given context is appropriate and aligned with foundational active listening best practices. No further feedback will be provided if the response is appropriate. 
    \item \textbf{Goal and Alignment}. Unlike casual conversations, peer counseling is goal-oriented, with each question or statement purpose-driven. This component defines what the counselor's goal in this part of the conversation should be and how the response can be changed to improve the alignment to this goal.
    \item \textbf{Areas for Improvement}. Re-iterating with domain experts and consulting mental health literature \cite{hill2009helping, 7cups2023},  we identify eight widely-used categories of effective communication for peer counseling context: \textit{Reflections, Questions, Suggestions, Validation, Self-disclosure, Empathy, Professionalism, Structure}. Areas of improvement highlights a set of categories that counselors need to further improve. 
    \item \textbf{Alternative Goal-Aligned Response} suggests an alternative response that aligns with the predefined goals and improves over these highlighted areas that need improvement, for a given context.
    \item \textbf{Positive Reinforcement} (optional) highlights a set of concrete categories as defined in \textit{Areas for Improvement} the peer counselors excel at. 
\end{enumerate}

Our multi-level feedback taxonomy, co-designed with senior psychotherapy supervisors, is the first of its kind to resemble how supervisors deliver feedback to counselors post-session. 
Unlike previous methods that only did one or the other, it uniquely combines evaluating responses and suggesting alternatives. Furthermore, the goal and alignment is a unique component of the taxonomy which explains how to improve alignment to a session-level goal.

\section{\textsl{FeedbackESConv} Dataset}
\label{sec:data}
In order to develop an automatic model that provides contextualized feedback at multiple levels, we use the feedback taxonomy to annotate peer counseling conversations. Given the sensitive nature of peer counseling data and the involved ethical implications, we chose a publicly available counseling dataset \emph{ESConv}  \cite{liu-etal-2021-towards} as our starting point, which contains a large number of emotional support conversations. ESConv was collected on a crowd-sourcing platform, thus requiring quality control. We performed a manual review to filter out conversations that were either low quality or irrelevant to peer counseling (refer to Appendix \ref{ESConv_filtering} for the comprehensive filtering criteria). We divided the obtained dataset into three parts: a dataset with 400 conversations for further annotation by domain experts; a dataset of 150 conversations (Preferences QESconv) used for obtaining self-scored preference pairs as described in Section \ref{sec:enhancement}; and a test dataset of 67 conversations.

\subsection{Domain Experts}
To obtain high-quality annotation, we take a user-centered approach by working with domain experts who have mental health expertise and hand-on practice experience.
We recruited domain experts from the Upwork platform by using a selective hiring process (see Appendix \ref{sec:appendix_hiring} for the hiring criteria). 
Our final annotator group consisted of two experts -- both with over 10 years of experience in professional mental health practice (one was a \textit{Certified Chemical Dependency Counselor} and the other an \textit{Associate Professional Clinical Counselor}).

\begin{table}[t]
\centering
\small
\begin{tabular}{lrr}
\hline
\multicolumn{3}{c}{\begin{tabular}[c]{@{}c@{}}  \textbf{FeedbackESConv}   \end{tabular}}  \\
\hline
Number of sessions & 400  \\
Number of utterances & 8179  \\
Number of appropriate utterances & 4721 & (57.7\%) \\
Number of inappropriate utterances & 3458 & (42.3\%) \\
Avg. length of alternative response & 28.3  \\
Avg. length of goal alignment & 36.6  \\

\hline
\textbf{Categories} & - & + \\
Reflections & 616 & 831 \\
Questions & 1431 & 1995 \\
Suggestions & 1159 & 259 \\
Validation & 901 & 1774 \\
Self-disclosure & 558 & 614 \\
Empathy & 1185 & 3313 \\
Professionalism & 279 & 462 \\
Structure & 333 & 1030 \\
\hline
\end{tabular}
\caption{FeedbackESConv: Statistics describing the number and average length of feedback annotations at different levels, as well the breakdown of highlighted categories for Areas of Improvement (-) and Positive Reinforcement (+).}
\label{tab:FeedbackESConv}
\end{table}

\subsection{Model-in-the-loop Co-annotation} \label{sec:data_annotation}
Recent work has shown that LLMs can offer a certain amount of facilitation for data annotation \cite{li-etal-2023-coannotating}. 
Thus, to facilitate the annotation process, we leverage a \emph{model-in-the-loop annotation paradigm}, with GPT-4 and domain experts working together on the annotation task -- the approach we later refer to as GPT-4+Expert. 

Before doing so, we rigorously compare the effectiveness of this co-annotation paradigm, where we set up a comparison of two approaches: generation of initial pre-annotations by GPT-4 and the subsequent refinement by experts, and annotations solely produced by experts. A full GPT-4 based annotation was technically possible, however, it was impossible to ensure feedback correctness and relevance without human supervision.

Our results (see Appendix \ref{sec:appendix_Expert_vs_GPT4+Expert}) show that in 80.8\% of cases, feedback created with GPT-4 pre-annotations is either preferred by experts (61.1\%) or there is no strong preference either way (19.7\%). This demonstrates the domain expert's preference for the model-in-the-loop co-annotation paradigm. 
As a result, during the annotation process, we use GPT-4 for the initial feedback annotation and then ask our experts to re-work these annotations. 

We prompt (see Appendix \ref{sec:appendix_prompt}) GPT-4 with detailed definitions of each of the feedback components (defined in Section \ref{sec:feedback_components}) and provide in-context examples containing feedback discussed with senior psychotherapy supervisors.
We provided domain experts with a detailed annotation guide (see Appendix \ref{sec:appendix_annotation}) with definitions and examples of each feedback component as described in our multi-level feedback taxonomy, to get them familiar with the task. 

This co-annotation produces 
annotations of over 400 emotional support conversations. We provide the detailed dataset statistics with the breakdown of highlighted categories for Areas of Improvement (-) and Positive Reinforcement (+) in Table \ref{tab:FeedbackESConv}.

\section{Model}
\label{sec:model}
\label{sec:enhancement}
We leverage the resulting FeedbackESConv dataset to develop models that can generate contextualized feedback at different levels for peer counseling. 
To enable transparent model development, we build upon the open-source Llama-2 model and introduce a simple but effective self-improvement method to generate multi-level feedback, as described below.

\subsection{Problem Definition}
\label{sec:problemdefinition}
Formally, we define the task of feedback generation based on our multi-level feedback framework as: (1) given the peer counselor's utterance $U_{i}$ and a context of the peer counselor-seeker conversation $c_{i}$, decide if the peer counselor's response is appropriate or needs further improvement by setting $y_{i}$ to $\mathrm{true}$ or $\mathrm{false}$, respectively. (2) If the response is classified as needing improvement, provide goal and alignment $goal_i$ (text), areas for improvement $ar-_i$ (list) and an alternative goal-aligned response $A_i$ (text). (3) Optionally, provide positive reinforcement or good areas $ar+_i$ (list) for this utterance as a form of positive reinforcement. We represent the feedback generation model as $\mathcal{M}$.

\subsection{Self-improvement via Forecasting}
\label{sec:technical_innovation}
\begin{figure}[t]
\centering
\includegraphics[width=2.6in]{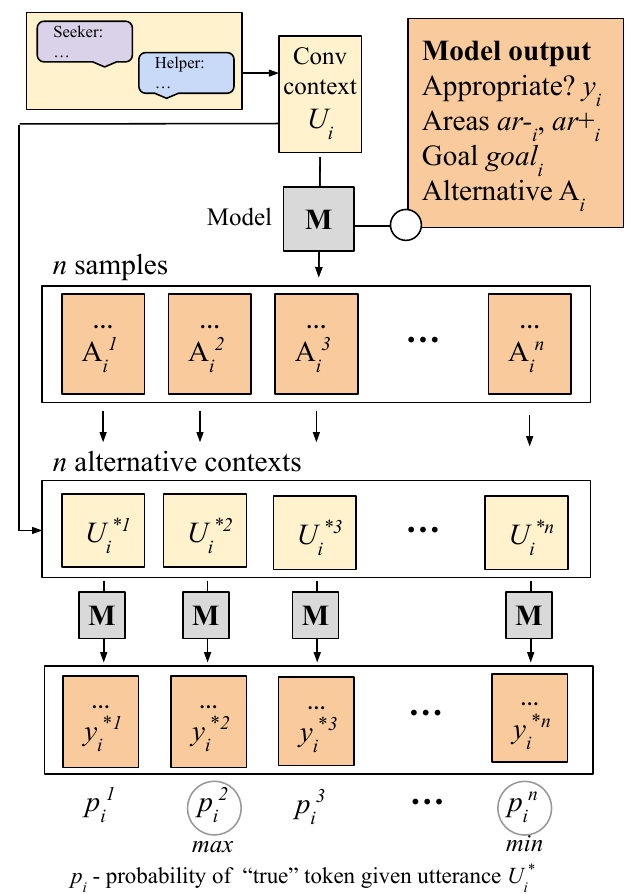}
\caption{Illustration of the self-scoring mechanism -- Phase 1 of the self-improvement method. 
The first step is to generate $n$ alternative answers for a given conversation utterance $U_i$. By substituting an alternative answer for the original utterance and passing it back to the model we obtain the probability of the alternative answer being marked as \textit{appropriate}. These scores can be used to create preference pairs for further alignment.}
\label{fig:selfperfecting}
\end{figure}

The specifics of our multi-level feedback framework allow us to suggest a self-improvement method for $\mathcal{M}$ that does not require any teacher model or additional costly expert data annotation.

On a high level, we take advantage of the fact that both response quality assessment ($y$) and alternative answer ($A$) are part of our feedback taxonomy. 
By substituting an alternative answer for the original utterance, our method uses the feedback model once again to forecast how generated alternative answers will be assessed.
This forecast operation estimates the quality of the originally generated feedback and can then be used to guide further alignment of the model. 
This self-improvement method has the potential to generalize to other scenarios since it applies to any model that jointly assigns binary $y_{i}$ ($\mathrm{false}$ or $\mathrm{true}$) label and suggests improvements for $y_{i} = \mathrm{false}$.

Concretely, to enable the self-improvement method with forecasting, we create self-scored preference pairs of feedback generations. To achieve that, we first establish a self-scoring function (Phase 1) and then use sample generations to choose the ones with maximum and minimum scores to form a pair (Phase 2). The model is then aligned to those self-scored preferences (Phase 3). 

\paragraph{Phase 1: Self-scoring} The goal is to establish a self-scoring function $\sigma$. Our feedback framework is designed in such a way that an alternative answer $A_{i}$ is part of the output of the model $\mathcal{M}(U_i)$. Hence, we can feed back the alternative answer $A_i$ to the original utterance $U_i$ and substitute it for the originally provided answer and obtain $U_i^{*}$ (Figure \ref{fig:selfperfecting}). This constitutes a self-assessment loop  because we can evaluate the quality of $U_i^{*}$ by once again passing it to $\mathcal{M}$. The proposed score is the probability of obtaining feedback labeled as \textit{appropriate} ($y_i = true$) for the refined utterance $U_i^{*}$. In summary, a feedback generation is assigned a high score if after following the advice (i.e. modifying the peer counselor's response in the suggested way) the probability of $y_i = true$ is high for this altered context.
This self-scoring mechanism is a proxy of feedback quality, as we assume that good feedback will lead to good alternative answers. 

\paragraph{Phase 2: Preference Pairs} Building on the self-scoring mechanism from Phase 1, these self-scores are obtained for a set of samples of $\mathcal{M}$ for the same utterance $U_i$. Samples with the maximum and minimum scores are indexed with $\omega_i$ and $\alpha_i$, respectively. 
If the probability that the original utterance $U_i$ receives feedback labeled as \textit{appropriate} is below 0.5 (indicating that further improvement is required), a preference pair is formed using samples $\omega_i$ and $\alpha_i$.

As a robustness check to assess whether these preference pairs are aligned with human judgment, we asked domain experts to annotate 20 test conversations with minimum and maximum score samples. They preferred the utterance with the higher score $63.0\%$ of the time, had no preference $28.9\%$, and only preferred the utterance with the lower score 8.1\% of the time.

\begin{figure*}[t]
\centering

\begin{tabular}{|c|c|c|c|c|}
\hline
Method & $\mathcal{M_{\text{SFT}}}$ & $\mathcal{M_{\text{SFT}}}$ + new data & $\mathcal{M_{\text{SFT}}}$ + best scores & $\mathcal{M_{\text{self-improvement}}}$\\ \hline
\makecell{Mean Score Overall} & 0.968 & 0.967 & 0.971 & \textbf{0.983*} \\ \hline
\makecell{Mean Score Worst 1\%} & 0.28 & 0.28 & 0.38* & \textbf{0.56*} \\ \hline
\makecell{Mean Score Worst 5\%} & 0.64 & 0.64 & 0.69* & \textbf{0.81*} \\ \hline
\end{tabular}

\includegraphics[trim={0cm 2.8cm 0cm 2.8cm}, clip,width=\textwidth]{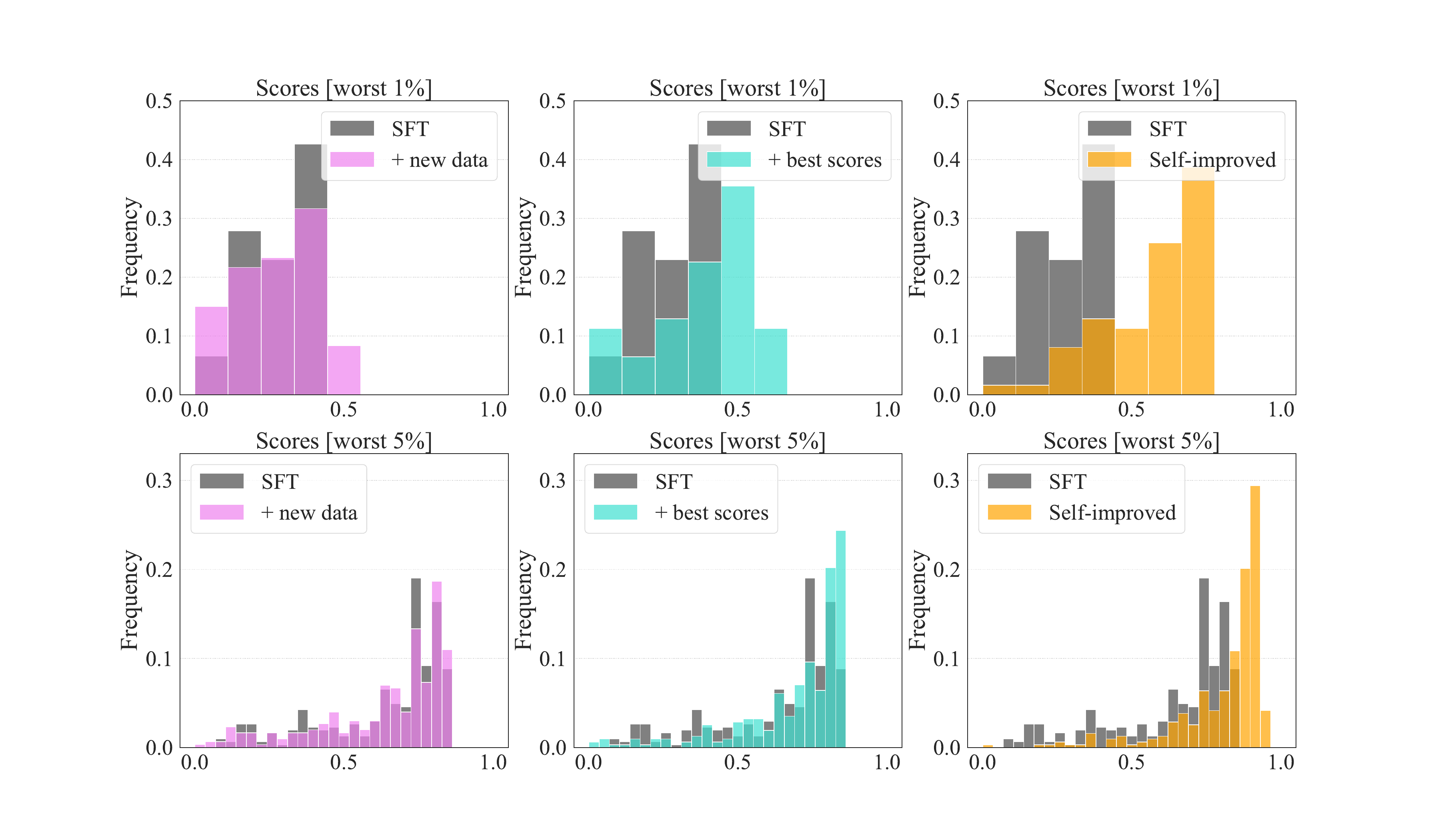}
\caption{Baselines comparisons. Table presents means of automatically-computed quality scores (as defined in Section \ref{sec:technical_innovation}) for three baselines and the self-improvement method. The comparison is shown for three different groups: overall and for the worst 1\% and 5\% of the generations. * denotes statistically significant (p < 0.01) improvements over  the $\mathcal{M_{\text{SFT}}}$ baseline based on t-test and Mann–Whitney U test. Plots present score distributions.}
\label{fig:scores_overall}
\end{figure*}

\paragraph{Phase 3: Alignment} The last step is to further align the model with Direct Preference Optimization (DPO) \cite{rafailov2023direct} to the preference pairs obtained from Phase 2. This technique contrasts high and low quality generations and encourages the model to produce generations similar to the ones marked as preferred. We align $\mathcal{M}$ on the Preferences QESconv dataset introduced in Section \ref{sec:data}. The resulting model is $\mathcal{M}_{\text{Self-imp}}$.

\subsection{Baselines}
\textit{$\mathcal{M_{\text{SFT}}}$ baseline}\footnote{Training details can be found in Appendix \ref{sec:appendix_experimental_setup}.}. 
To evaluate the self-improvement via forecasting method, we compare it with a supervised fine-tuned Llama2 13B model baseline, denoted as $\mathcal{M_{\text{SFT}}}$. 
To understand whether the different phases in the self-improvement method are essential, we compare it with two additional baseline ablation conditions:

 \textit{$\mathcal{M_{\text{SFT}}}$ + new data}.  We apply the $\mathcal{M_{\text{SFT}}}$ model to obtain feedback generations for the additional data Preferences QESConv that $\mathcal{M_{\text{Self-imp}}}$ uses.
 We use those generations for further supervised fine-tuning. The goal here is to determine if self-scoring gives value beyond simply fine-tuning on additional generations on new data used by $\mathcal{M_{\text{Self-imp}}}$.

\textit{$\mathcal{M_{\text{SFT}}}$ + best scores}. We follow the self-scoring procedure, but instead of creating a single preference pair, we generate multiple scored samples and choose the one with the highest score for further fine-tuning the $\mathcal{M_{\text{SFT}}}$ model. The aim is to see whether alignment to preference pairs gives improvement compared to fine-tuning to the highest scored generation.

\section{Evaluation and Results}

\label{sec:evaluation}
\label{sec:evaluation_results}

In Section \ref{sec:minimizing_risk}, we compare the quality of feedback generated with $\mathcal{M}_{\text{Self-imp}}$ vs. those generated with baseline models via automatic scores and domain-expert ratings.  
After validating the improved feedback quality of $\mathcal{M}_{\text{Self-imp}}$ over baselines, in Section \ref{sec:overall_feedback_quality}, we compare its feedback to the feedback co-annotated by GPT-4+Experts (approach described in Section \ref{sec:data_annotation}) to understand if the $\mathcal{M}_{\text{Self-imp}}$ model matches in quality.
\if 0

\fi

\subsection{Comparing $\mathcal{M}_{\text{Self-imp}}$ with Baselines} 
\label{sec:minimizing_risk}
We use the automatically-computed quality scores (as defined in Section \ref{sec:technical_innovation}) as one way to evaluate the performance of our self-improvement method against baselines\footnote{To ensure a fair comparison, we perform scoring using the same base $\mathcal{M_{\text{SFT}}}$ model.}.
For each model, we generate 10 samples of feedback for each counselor utterance in 67 test conversations resulting in 8090 data points.
Our results are reported in Figure \ref{fig:scores_overall}.
Over all feedback generations, the mean quality score is highest for $\mathcal{M}_{\text{Self-imp}}$, where the difference compared to $\mathcal{M}_{\text{SFT}}$ is statistically significant.  

\begin{figure}[t]
\centering
\includegraphics[width=0.48\textwidth]{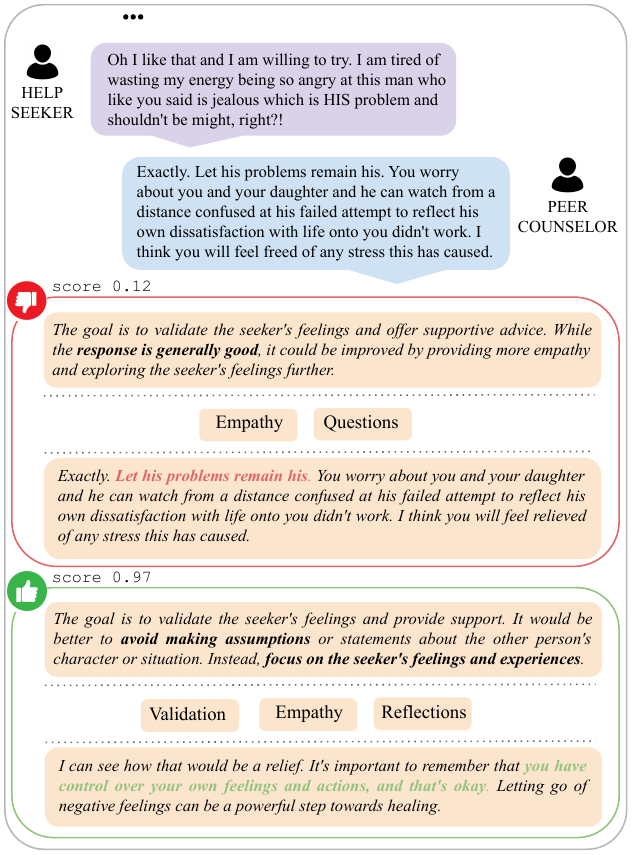}
\caption{Example of feedback response of very poor quality. The model incorrectly provided feedback that the peer counselor response is \textit{generally good}. Although the model properly outlined the intended goal of peer's counselor reply, the proposed alternative fails to align with this goal and repeats the same errors. A representation of what constitutes high-quality feedback generation for this specific instance is provided for clarity.}
\label{fig:qualitative_example}
\end{figure}

In the context of peer counseling, unlike typical natural language generation tasks where average performance is key, our focus is on minimizing the chance of producing poor or unhelpful feedback, prioritizing the worst-case scenario. We illustrate this with an example of both low quality and high quality feedback in Figure \ref{fig:qualitative_example}.

As shown in the table in Figure \ref{fig:scores_overall}, in the worst 5\% and 1\% of generated feedback, the quality scores for the $\mathcal{M}_{\text{Self-imp}}$ model are significantly higher than the baselines. For the bottom 1\% of samples, the mean score increases from 0.28 for $\mathcal{M}_{\text{SFT}}$ to 0.56 for $\mathcal{M}_{\text{Self-imp}}$, indicating a reasonable shift from inappropriate to appropriate feedback.

\begin{figure}[t]
\centering
\includegraphics[trim={0cm 0.73cm 0cm 1.4cm}, clip,width=0.48\textwidth]{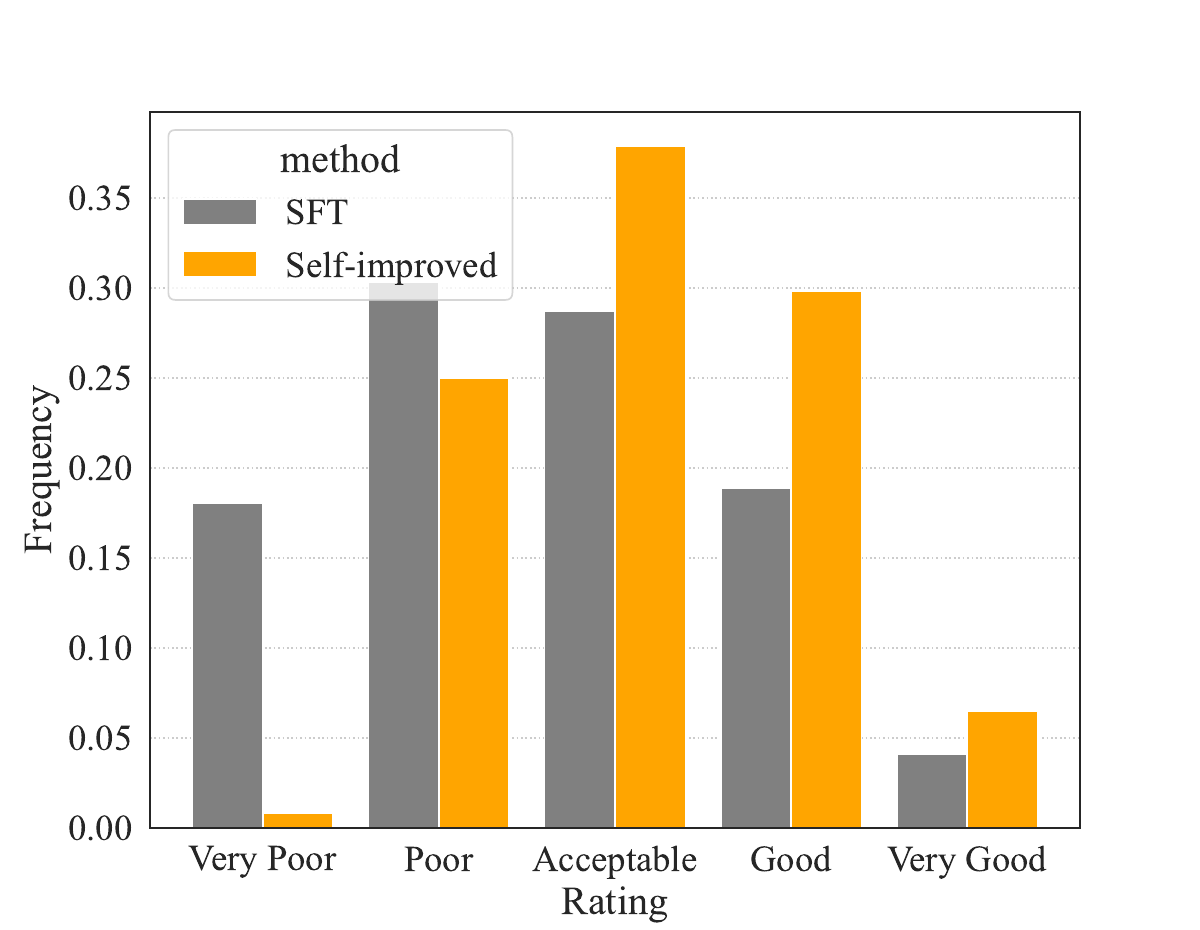}
\caption{Expert quality assessments for the worst 1\% of generations. 
The statistically significant shift of scores to the right ($p$ < 0.01) shows the \textit{self-improvement} method was judged to be of higher quality than the $\mathcal{M}_{\text{SFT}}$ baseline, with mean score improving from 2.61 (Below Acceptable) to 3.16 (Above Acceptable).}
\label{fig:human_eval_all}
\end{figure}

Automatically-computed quality scores enable observations of improvements on the aggregate distribution level. To affirm that our proposed method $\mathcal{M}_{\text{Self-imp}}$ enhances the quality of feedback in the worst-case scenario, we defer to the gold standard of evaluation: the judgment of domain experts.

We conducted the following experiment. We asked domain experts to rate the feedback quality of the bottom 1\% of generations using a 5-point Likert scale for $\mathcal{M}_{\text{SFT}}$ and $\mathcal{M}_{\text{Self-imp}}$. As shown in the bottom of Figure \ref{fig:human_eval_all}, generations rated as \textit{Very Poor} were almost all eliminated by the use of the $\mathcal{M}_{\text{Self-imp}}$ method, to less than 1\% of the ratings. Moreover, we see consistent growth of the proportion of generations marked as \textit{Acceptable}, \textit{Good} or \textit{Very Good}. One author further conducted a qualitative investigation of the worst 1\% of feedback. We observe that feedback from $\mathcal{M}_{\text{SFT}}$ can often suggest alternative answers with slight rephrasing that do not resolve the core issue, whereas 
 $\mathcal{M}_{\text{Self-imp}}$ exhibits fewer of these errors.    

Together, the results from these two experiments suggest that for the worst generations, $\mathcal{M}_{\text{Self-imp}}$ improves feedback quality as measured both by automatically-computed quality scores and domain expert ratings.

\subsection{Comparing $\mathcal{M}_{\text{Self-imp}}$ with GPT-4+Expert}
\label{sec:overall_feedback_quality}
% Other names: Overall Utility Evaluation

\begin{table}[t]
\centering

\resizebox{0.48\textwidth}{!}{
\begin{tabular}{lcc}
\hline
\textbf{Feedback Aspect} & \textbf{$\mathcal{M_{\text{self-imp}}}$} & \textbf{GPT-4 + Expert} \\
\hline
Selection for Feedback & 4.20 & 4.18 \\
Strengths Identification & 3.68 & 3.95 \\
Improvement Areas Selection & 4.28 & 4.3 \\
Goal Description Quality & 4.3 & 4.43 \\
Rationale for Alternatives & 4.33 & 4.45 \\
Quality of Alternatives & 4.03 & 4.38$^{*}$ \\
Feedback Style & 4.45 & 4.55 \\
Feedback Helpfulness & 4.15 & 4.48$^{*}$ \\
\hline
Overall & 4.10 & 4.35$^{*}$ \\
\hline
\end{tabular}
}
\caption{Experts' conversation level evaluation of eight aspects of feedback quality for $\mathcal{M_{\text{Self-imp}}}$ and the reference GPT-4+Expert annotations. Results based on a test sample of 20  conversations. * denotes statistically significant difference under t-test ($p< 0.05$).}
\label{table:comparison}
\end{table}

We further assessed feedback quality at the conversation level and compared feedback generated by $\mathcal{M}_{\text{Self-imp}}$, against the GPT-4+Expert annotations. Domain-experts evaluated the quality of feedback along eight aspects that cover the components of the multi-level feedback taxonomy. Results (Table \ref{table:comparison}) indicate that the $\mathcal{M}_{\text{Self-imp}}$ model's feedback quality approaches the reference standard of GPT-4+Expert annotations across 6 out of 8 feedback aspects, with a median overall quality rating of \emph{4 - Good}.  
We note significant differences in the \emph{Quality of Alternatives} and overall \emph{Feedback Helpfulness}.
Nevertheless, we find that experts agree (4 or 5 on the Likert-scale) in 90\% of conversations that the feedback generated by $\mathcal{M}_{\text{Self-imp}}$ would be helpful in the training process of novice peer counselors (100 \% of GPT-4 + Experts annotations are considered helpful).
These results validate how $\mathcal{M}_{\text{Self-imp}}$, a model based on Llama-13B trained using our self-improvement method, can match the GPT-4+expert reference annotations across many aspects, while highlighting aspects of the multi-level feedback taxonomy that future modeling work can improve. 
Example feedback generations are in Appendix \ref{sec:appendix_generations}.

\section{Conclusions}
\label{sec:conclusions}
We introduced a multi-level feedback framework for training counseling skills by co-designing with senior psychotherapy supervisors, constructed a public dataset of counseling conversation with feedback annotations, and  proposed a simple but effective self-improvement method for feedback generation. We demonstrate through qualitative and quantitative evaluation that our method minimizes the risk of low-quality feedback generation and generates feedback that domain experts find useful. 

\section*{Limitations}
In this work, we first co-designed with senior psychotherapy supervisors a feedback framework and then developed a LLM model that can automatically generate advice for novice peer counselors. Although the framework covers multiple aspects of active listening, it is not enumerative and might not cover all possible feedback dimensions  relevant to the complex peer counseling context.

While we consider the way in which the feedback is delivered (and specifically evaluate the feedback style -- whether it was delivered "in a friendly but professional way"), we do not tailor our feedback to a specific trainee in a personalized way. In professional training of therapists, supervisors alter their feedback style to optimize feedback delivery: \textit{``But in addition I have a take on who is this person I'm supervising. And what are they like as a person? And do they listen to me or not? And how can I say it differently so they can hear it?''}

Our feedback dataset, which we used for training of our model, was built on a public dataset of emotional support conversations. This allows us to make our data publicly available. However it was built upon conversations between crowd workers who have only received very abbreviated training. While the training covers a broad range of counseling skills, it is unclear whether these crowd-sourced conversations might generalize to conversations among peer counselors and seekers or other similar counseling contexts.

Although we involved human experts (senior psychotherapy supervisors and domain experts with counseling expertise) at every stage of the development process and system evaluation, we acknowledge that the opinions and judgments from this small group of domain experts might not represent a broader population of psychotherapy supervisors or mental health practitioners, as well as the ways in which they coach novice peer counselors.

\section*{Ethics Statement}
This study 
has been approved by the Institutional Review Boards (IRB) at the authors’ institutions.  All the researchers involved in this study have completed CITI Program certifications on responsible code of conduct in research.
We have compensated domain experts fairly for their time, going beyond minimum wage in the United States.  

The purpose of this paper is to develop a model that generates feedback for novice peer counselors with limited or no access to human supervision. The system should not be regarded as a substitute for expert feedback. 
Importantly, while our self-improvement method aims to limit the risk of poor feedback generations (e.g., giving inappropriate advice), this risk is not fully eliminated. It is therefore important to treat model-generated advice only as potential guidance and discard it if necessary, based on trainee judgment.

For potential uses of this feedback generation system, we will design consent form to disclose potential risks of our system, and will also advocate for practitioners to centrally host and log the content generated by our system so that it can be audited to determine whether there are any problematic behaviors in the system use.

\bibliography{anthology,custom}
\bibliographystyle{acl_natbib}

\newpage
\appendix

\section{Evaluation areas}
\label{sec:appendix_areas}

Table \ref{tab:effective_communication} presents specific example mistakes which peer counselors can make. These are grouped into 8 categories with definitions aligned with mental health literature. 
\begin{table*}[ht]
\centering
\resizebox{\textwidth}{!}{
\begin{tabular}{rp{1.2\textwidth}}
\hline
\hline
\textcolor{myColour}{Reflections} &  This skill involves repeating or rephrasing clients' statements to identify and acknowledge their feelings. This technique helps clarify the client's emotions and encourages them to explore these feelings further. \\
\hline
\textbf{References:}&  \cite{bugental2001handbook, rautalinko2007reflective, arnold2014behind, hill2009helping, moyers2014motivational, moyers2016motivational, perez-rosas-etal-2019-makes, beck2020cognitive, shah2022modeling} \\
\hline
\textbf{Example mistakes:}& \small  Not reflecting, drawing conclusions from the helper's experience without listening to what the seeker is saying and checking it out with them; Making assumptions beyond what was said; Copying the seeker's words exactly; Stating feelings too definitely rather than tentatively (e.g., "you obviously feel X" vs. "I wonder if you feel X"); Becoming repetitive, not varying the format of restatements (e.g., I’m hearing you feel sad, I’m hearing you have some thoughts about X, I’m hearing you ...); Labeling feelings inaccurately; Not capturing the most salient feeling; Reflecting on many feelings at the same time; Being judgmental; Focusing on the feelings of others and not the seeker; Reflecting when the seeker is resistant to expressing feelings and reflection might add more pressure. \\
\hline
\hline
\textcolor{myColour}{Questions} &  Questions in peer counseling can be formulated either as inquiries (e.g., "How do you feel about that?") or as prompts (e.g., "Tell me more about your feelings on that"), provided to aid the client in understanding or examining their emotions. \\
\hline
\textbf{References:} &  \cite{bugental2001handbook, hill2009helping, james2010science, moyers2014motivational, moyers2016motivational, beck2020cognitive, shah2022modeling} \\
\hline
\textbf{Example mistakes:} & \small  Making questions too focused in situations in which they should be more open-ended; Trying to cover everything instead of focusing on one aspect; Asking questions without a clear intention/goal; Not encouraging expression of feelings; Not exploring the details of the situation the seeker is coming with; Not asking the seeker to check the facts ("tell me what data you have that supports that", "do you have any evidence that you'd be X if you did Y?"); Asking questions without empathy; Asking lengthy or multiple questions at once; Turning the attention to other people instead of the seeker (i.e., asking what person X did, instead of asking how the seeker felt about X’s behavior); Asking too many closed-questions interviewing instead of exploring. \\
\hline
\hline
\textcolor{myColour}{Suggestions} &  This technique involves offering specific directives or advice that clients can apply outside the counseling sessions. \\
\hline
\textbf{References:} &  \cite{bugental2001handbook, hill2009helping, moyers2014motivational, moyers2016motivational, beck2020cognitive, shah2022modeling} \\
\hline
\textbf{Example mistakes:} & \small  Giving too much or premature advice, answers, or solutions; Telling people what to do, giving direct advice "you should"; Imposing beliefs or personal values on seekers; Trying to debate with the seeker and convince them of the helper's point of view.
\\
\hline
\hline
\textcolor{myColour}{Validation} &  Validation goes beyond simply acknowledging a client's feelings. It actively affirms their experiences and perspectives as understandable and worthy of respect, even if the counselor may not personally share their viewpoints. \\
\hline
\textbf{References:} &  \cite{linehan1997validation, bugental2001handbook, hill2009helping, moyers2014motivational, moyers2016motivational, beck2020cognitive} \\
\hline
\textbf{Example mistakes:} & \small  Not letting the seeker know that their feelings are normal; Validating invalid (e.g., validating opinions or seeker’s biases); Helper not being there, paying attention to what the seeker brings to the conversation.
\\
\hline
\hline
\textcolor{myColour}{Self-disclosure} &  Sharing of personal experiences can create a sense of empathy and connection, reducing the client's feeling of isolation. This approach is balanced to avoid overshadowing the client's emotions or introducing irrelevant personal details. \\
\hline
\textbf{References:} &  \cite{henretty2010role, bugental2001handbook, hill2009helping, moyers2014motivational, moyers2016motivational, beck2020cognitive, shah2022modeling} \\
\hline
\textbf{Example mistakes:} & \small  Not turning the focus back to the seeker immediately; Making self-disclosure too long or too complex; Disclosing too much information; Talking too much and not letting the seeker talk more. \\
\hline
\hline
\textcolor{myColour}{Empathy} &  This skill involves understanding the client's emotions and sharing in their experience, offering a sense of being truly seen and heard. This deeper connection allows counselors to guide clients toward self-discovery and provide targeted support. \\
\hline
\textbf{References:} &  \cite{bugental2001handbook, hill2009helping, beck2020cognitive, cooper2020mindfulness, sharma-etal-2020-computational} \\
\hline
\textbf{Example mistakes:} & \small  [Empathetic Emotional Reactions] Not expressing warmth, compassion, concern, or similar feelings towards the seeker in situations in which it would be appropriate; [Empathetic Interpretations] Not communicating an understanding of the seeker's experiences and feelings in situations in which it would be appropriate; [Empathetic Explorations] Not making an attempt to explore the seeker's experiences and feelings in situations in which it would be appropriate; Expressing empathy but without maintaining a professional attitude; Expressing sympathy instead of empathy. \\
\hline
\hline
\textcolor{myColour}{Professionalism} &  Professionalism refers to setting clear boundaries and using appropriate language and communication style. \\
\hline
\textbf{References:} &  \cite{bugental2001handbook, hill2009helping} \\
\hline
\textbf{Example mistakes:} & \small  Overusing slang; Being overly professional and formal, which results in robotic-style conversations; Using vocabulary that expresses too much closeness. \\
\hline
\hline
\textcolor{myColour}{Structure} &  This skill assists the counselor and client in guiding the conversation effectively, ensuring productive use of time, and covering essential topics. A basic structure, while flexible to individual needs, provides both parties with a sense of security and direction.  \\
\hline
\textbf{References:} &  \cite{day1980structuring, bugental2001handbook, hill2009helping, moyers2014motivational, moyers2016motivational, beck2020cognitive} \\
\hline
\textbf{Example mistakes:} & \small  [beginning] Not establishing a collaborative agenda and a friendly emotional rapport; [middle] Having too many topics on the table at the same time, not focusing on the main problem ("keep it simple"); [end] Not summarizing what the person is going to take away from the conversation; [end] Lack of clear, actionable items or insights for the seeker after the conversation. \\
\hline
\hline
\end{tabular}
}
\caption{Examples of evaluation areas in peer counseling communication grouped into 8 categories: \textit{Reflections, Questions, Suggestions, Validation, Self-disclosure, Empathy, Professionalism, Structure}.}
\label{tab:effective_communication}
\end{table*}

\section{Interviews with senior experts}
\label{sec:appendix_interview}

To understand the nature of feedback in professional training, we conducted multiple interviews with three senior psychotherapists with over 20 years of direct supervision experience of novice therapists. We first understood the common practices of feedback-giving sessions and then engaged with supervisors on a representative task of providing feedback on a transcript of an emotional support conversation to simulate the process of communicating feedback to a psychotherapist student.

The interviews focus on the following questions, insights from which guided the framework design process:
\begin{itemize}
\itemsep0em 
    \item[--] R1: What are important skills for novice counselors? 
    \item[--] R2: How are these skills learned and what is the role of feedback in the learning process?
    \item[--] R3: What is the structure of this feedback?
\end{itemize}

We transcribed all audio recordings of the interviews. Then, using a thematic coding \cite{terry2017thematic} approach, we analyzed the interview transcripts to identify key themes and patterns across the data. We then studied how those inform our research questions.

\subsubsection*{R1: What are important skills for novice counselors?}
Beginner psychotherapy skills involve increasing the depth of self-description of the support seeker's problems. Experienced psychotherapists can perceive nuances and undertones in conversations that beginners might miss. 
\begin{displayquote}
[Supervisor 1] \textit{"I think an experienced psychotherapist can hear some, can hear some things or pick up on some things that a novice therapist maybe won't that are between the lines."}
\end{displayquote}
The main objective for beginners is not necessarily about adhering to a particular model but mastering basic foundational skills. 
\begin{displayquote}
\textit{"I'm thinking that with the beginning novice therapist it's less the model than sort of basic foundational skills that we, I think, we're trying to teach"}
\end{displayquote}
Our experts often referred to \textit{“Helping Skills: Facilitating Exploration, Insight, and Action”} textbook by Carla Hill, who has devised a system categorizing these essential helping skills. 

The initial training phase focuses on foundational listening skills, which are also crucial for peer counselors to master.

\subsubsection*{R2: How are these skills learned and what is the role of feedback in the learning process?}
Early-stage students undergo training in foundational counseling skills like listening, empathy, and asking open-ended questions. 

\begin{displayquote}
[Supervisor 1] \textit{"Students in the beginning, they, they take certain classes on what I might call basic foundational counseling skills, how to listen, how to be empathetic, how to, you know, ask open-ended questions. There's a list. You know, there's a list of skills"}
\end{displayquote}

After their first year, novice therapists undergo a practicum experience where their sessions are taped and reviewed for feedback on foundational skills.

\begin{displayquote}
[Supervisor 1] \textit{"At the end of their first year, they go for their first clinical experience. We call it practicum experience. And their sessions are taped, and their supervisor goes over those tapes with them and gives them feedback on their, you know, on how they're doing on those basic skills."}
\end{displayquote}

It's beneficial for novices to bring session transcripts, as these provide clear evidence of their actions and their consequences. These tapes and transcripts allow both the supervisor and the novice to study the impact of the therapist's actions on the patient.

\begin{displayquote}
[Supervisor 2] \textit{"But also I like them to bring a transcript. Because then I can go show them. See what you did here led to this, which led to this, and this is what you should do instead."}
\end{displayquote}

\subsubsection*{R3: What is the structure of this feedback?}
When providing feedback to the novice therapist, the experts emphasized the importance of positive reinforcement by starting with what the counselor did well. 

\begin{displayquote}
[Supervisor 2] \textit{"I generally start out with. What they're doing well"}
\end{displayquote}
\begin{displayquote}
[Supervisor 3] \textit{"Well, I'd say this is pretty good overall, so I'd give positive feedback first."}
\end{displayquote}

They would then gently introduce \textcolor{blue}{\textbf{areas for improvement}}. The two crucial skills are making proper reflections and asking good open-ended questions. However, many other areas were mentioned by the experts as they analyzed the provided conversation transcripts.

\begin{displayquote}
[Supervisor 3] \textit{"paraphrasing is a main thing. It's just a couple of, I think that's an important piece. You ask the person a question or they start and then you just kind of repeat what they say [...]  asking open questions is another really good one thing that people learn to do"}
\end{displayquote}

Using transcripts like the discussed one can be an effective teaching tool, prompting the therapist to think of \textcolor{blue}{\textbf{alternative responses}}.

\begin{displayquote}
[Supervisor 3] \textit{"I would teach it by using a transcript like this. And then I'd say [... ] what other kinds of things can you think of that if I said them to you, you'd be more likely to really sink into what it is you're trying to come and talk about?"}
\end{displayquote}

Counseling should be \textcolor{blue}{\textbf{goal-focused}}, each question or statement should have a goal.

\begin{displayquote}
[Supervisor 3] \textit{"[...] what were your goals right? What were your goals in making these questions or suggestions or statements? And I would have have them try and think about it."}
\end{displayquote}

When going back to the transcript, the expert analyzed it line by line, stopping at each of the helper’s responses and giving feedback on it.

\begin{displayquote}
[Supervisor 2] \textit{"Counselor says “she gave you a lot of meeting and filled your time fondly”. OK, So she's interpreting his statement rather than pulling out more of his statement."}
\end{displayquote}

Crucially, the experts point out that the delivery of feedback should be in a manner that ensures the counselor doesn't feel criticized.

\begin{displayquote}
[Supervisor 2] \textit{"How do they deliver it so that the therapist can hear it? And how does the therapist work with the patient? There are two communications going on there"}
\end{displayquote}

Senior supervisors were compensated \$150/hour.

\section{ESConv filtering}
\label{ESConv_filtering}

\begin{figure}[th]
\centering
\includegraphics[width=3in]{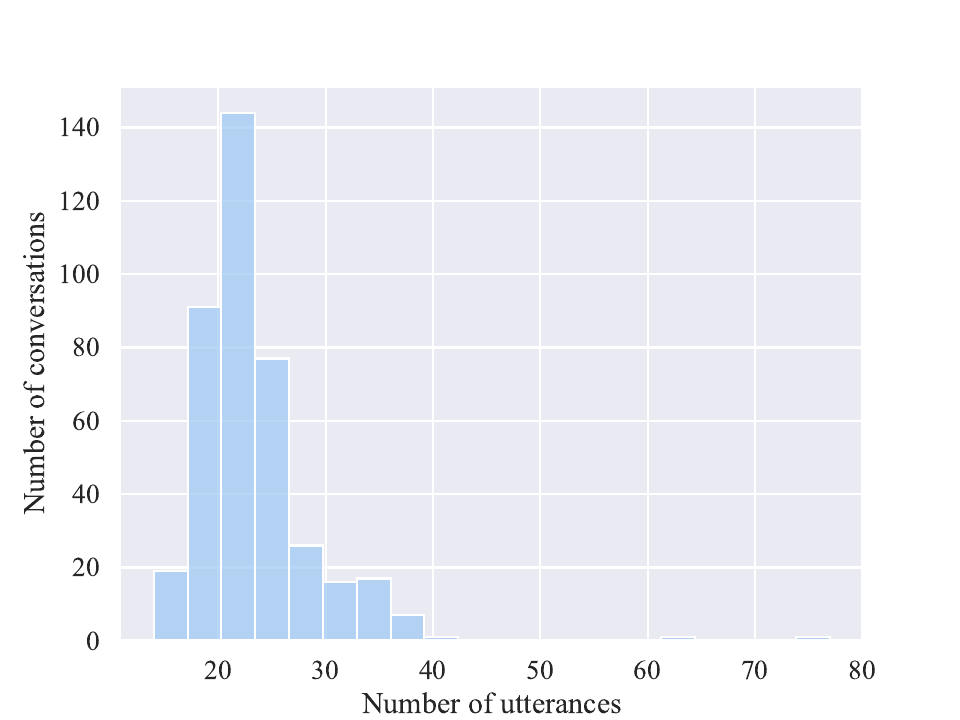}
\caption{QESConv distribution of the number of utterances in conversations.}
\label{fig:gesconv_utts}
\end{figure}

\begin{figure}[th]
\centering
\includegraphics[width=3in]{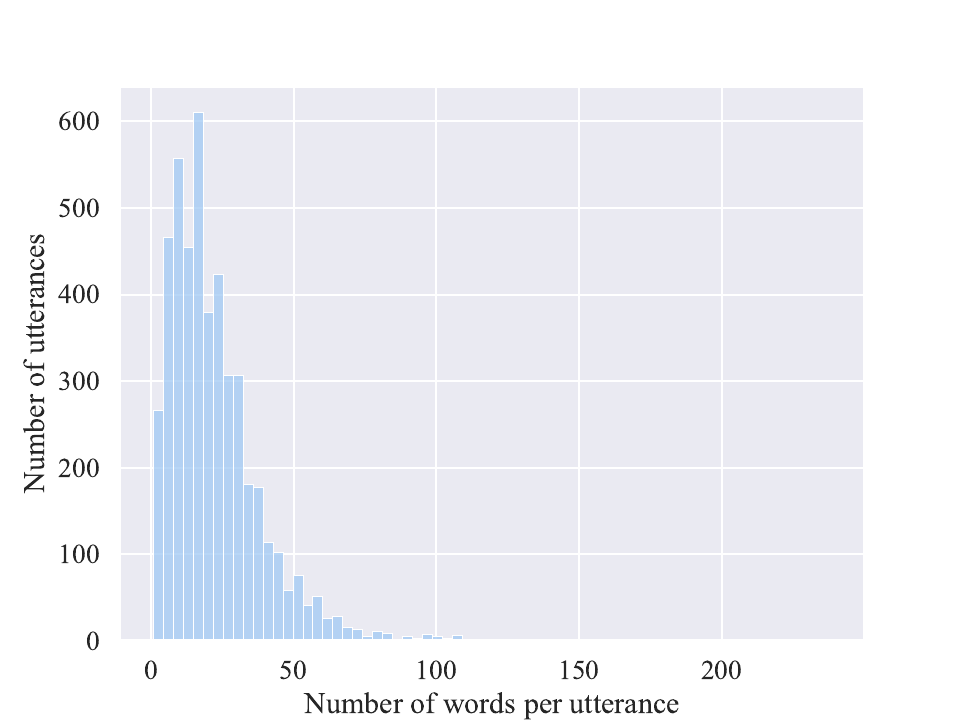}
\caption{QESConv distribution of the number of words in helper's utterances.}
\label{fig:gesconv_utts_helper}
\end{figure}

\begin{figure}[th]
\centering
\includegraphics[width=3in]{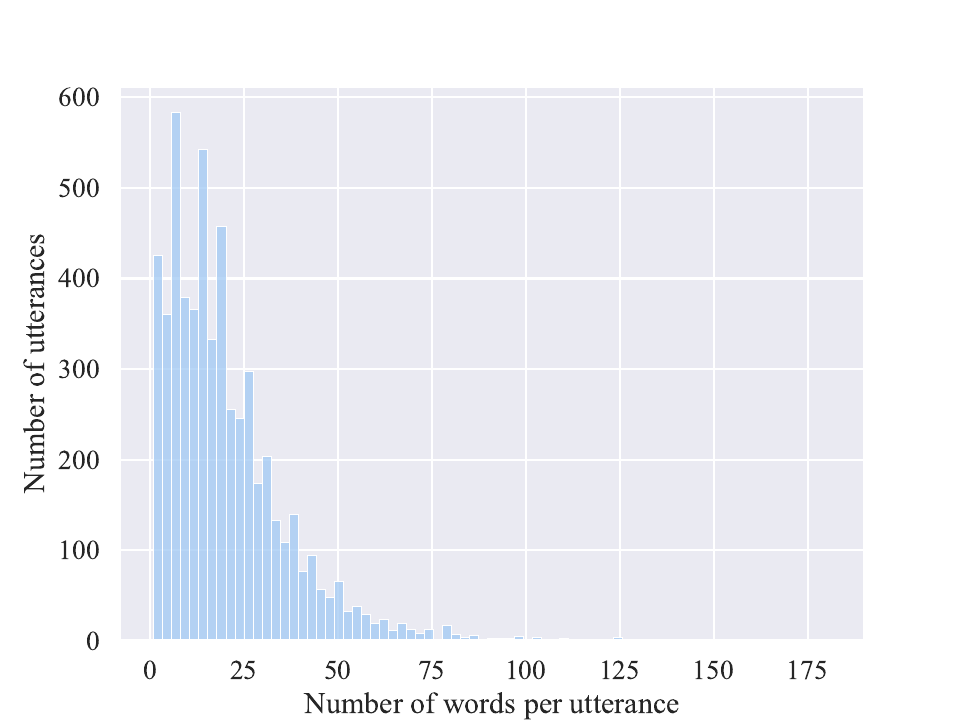}
\caption{QESConv distribution of the number of words in seeker's utterances.}
\label{fig:gesconv_utts_seeker}
\end{figure}

We manually analyze the  conversations in ESConv \cite{liu-etal-2021-towards} (CC BY-NC 4.0 license) and filter the ones that meet the following criteria:
\begin{itemize}
\itemsep0em 
    \item[--] Conversation not on topic
    \item[--] Conversation referring in big part to MTurk
    \item[--] Conversation not serious: making jokes, etc.
    \item[--] Ungrammatical
    \item[--] Chatting mostly about the current situation COVID, not a specific problem (i.e., exchanging news, vaccination discussions, etc.)
    \item[--] Mostly meta-conversation (``sorry, are you there, I have not seen your message'')
    \item[--] Generic topic chat: hobbies, having a dog, looking for job advice
\end{itemize}

In this way we select 400 conversations for the QESconv dataset.  
We further remove many conversation-finishing artifacts by searching for keywords ``survey'', ``quit'', ``we need to chat'',``button'' and manually removing those from utterances.
For example: \textit{``can you press quit first, I can't do it from my end'' }, \textit{``I think we need to chat a bit more in order to wrap things up''}, \textit{``please remember to take the survey :)''}, \textit{``Is there a quit/finish button on your end?''}. 

The final dataset has in total 11.3K utterances (distribution shown in Figure \ref{fig:gesconv_utts}, with average utterance length equal 21.4 words (distribution for helper in Figure \ref{fig:gesconv_utts_helper} and seeker in Figure \ref{fig:gesconv_utts_seeker}).

\section{Domain experts hiring process}
\label{sec:appendix_hiring}

Based on the submitted applications and conducted interviews, we choose a group of six experts. We then conduct a pilot study in which we ask the experts to annotate a single conversation based on our annotation guide describing the feedback framework (Section \ref{sec:feedback_framework}) and our annotating interface (Appendix \ref{sec:appendix_annotation}). Based on adherence to the guide and projected time availability, we establish a group of three self-validated (at least 4/5 in the Likert scale -- for details see Appendix \ref{sec:appendix_quality}) experts  -- all with over 10 years of professional mental health practical experience (for example as \textit{Certified Chemical Dependency Counselor}, \textit{Licensed Marriage and Family Therapist} and \textit{Associate Professional Clinical Counselor}). 

Upon further quality tests for the final data annotation scheme, we narrow down the group to two experts who consistently validate the quality of each other's annotations on the final annotation task (see Appendix \ref{sec:appendix_final_validation_annotators}). Our annotators are US-based.

Domain experts were compensated \$30/hour. We informed them of the purpose of the study and the potential risks.

\section{Pilot quality validation}
\label{sec:appendix_quality}

We observe variability in feedback among experts, but we confirm with senior supervisors that this is to be expected since each practitioner may focus on different counseling components.
Since there is no gold truth feedback, \footnote{Even identifying areas for improvement cannot be simply defined as a multi-classification problem since different areas can be highlighted and there is not a single correct set} evaluating the annotation quality is challenging and requires human expertise. We, therefore, perform a pilot self-validation study in which each expert judged  on a 5-point Likert scale the quality of of the other experts' annotations.

In an experiment involving three experts (third expert later excluded at the co-annotation stage), each was tasked with evaluating the annotations made by the others for a single conversation. The assessment was based on a five-point Likert scale:

\begin{enumerate}
    \item \textbf{Completely Irrelevant}: The feedback is unrelated to the task.
    \item \textbf{Slightly Relevant}: The feedback has minimal relevance, lacking depth or specificity.
    \item \textbf{Moderately Relevant}: The feedback is partially relevant, covering some, but not all, key aspects.
    \item \textbf{Highly Relevant}: The feedback addresses most key aspects effectively.
    \item \textbf{Exceptionally Relevant}: The feedback is comprehensive, insightful, and offers actionable suggestions.
\end{enumerate}

Even though the annotations varied, the experts found different ways of giving valid feedback: \textit{``I think the other annotators and I emphasized things in slightly different ways. For example, one was more focused on clarity and the other was more focused on validation.''} They all rated each others annotations to be at least 4/5 validating the overall annotation quality (see Table \ref{tab:pilot_validation}). 

All evaluations were blind, i.e. we did not reveal the source of the annotations.

\begin{table}[ht]

\centering
  \begin{tabular}{|c|c|c|c|}
  
    \hline
    \multirow{2}{*}{Evaluator} & \multicolumn{3}{c|}{Annotator} \\ \cline{2-4}
     & A & B & C \\ \hline
    Expert A & - & 4/5 & 4/5 \\ \hline
    Expert B & 5/5 & - & 4/5 \\ \hline
    Expert C & 4/5 & 5/5 & - \\ \hline
  \end{tabular}
  \caption{Quality validation pilot results.}
  \label{tab:pilot_validation}
\end{table}

\section{Potential of LLMs for providing feedback}
\label{sec:appendix_LLM_feedback_potential}

We explore whether LLMs could help in the annotation process within the feedback framework we have defined. This presents a topic of empirical investigation on its own. 

LLMs have been used for annotations \cite{gilardi2023chatgpt, kuzman2023chatgpt}, or co-annotation  \cite{li-etal-2023-coannotating}, and GPT-3.5 and GPT-4 models excel at classification tasks related to client/therapist behaviors \cite{chiu2024computational}; however, our task is much more open-ended, requiring a generation of natural language rationale using deep understanding of the specialized feedback framework.

We experiment with Llama2-70b chat \cite{touvron2023llama}, GPT-3.5 Turbo and GPT-4 models \cite{Achiam2023GPT4TR}. While all models give reasonable feedback when prompted with a short generic statement, \footnote{Example simple prompt: \textit{Act as a supervisor of novice helpers in the mental health context. Give feedback to the helper on their last response in the conversation below.}} the feedback is not focused (the most generic for the Llama model).

When provided with a detailed definition of our framework, we find Llama to be unsuccessful in parsing framework guidelines, which both GPT-3.5 and GPT-4 manage to do. However, we find in early experiments that GPT-3.5 produces feedback of significantly inferior quality to human one, therefore we proceed with GPT-4 as our base model, which showed high potential.

\subsection{GPT-4 prompting}
\label{sec:gpt4_model}

While the most straightforward approach would be to use an API call to annotate each $U_{i}$, it would be very expensive given the usage of the GPT-4 model and the number of tokens in the instruction (>2k). Annotating the full conversation at once would be the most efficient option, but we notice a significant degradation of quality in annotations of the final helper's utterances. Therefore, we annotate overlapping chunks for the conversation of 5 helper's utterances \footnote{The chunks are overlapping; we discard feedback for the first two utterances, which lack the sufficient context.}.

\subsection{GPT-4 quality pilot}
\label{sec:appendix_quality_GPT}

Similar to the setting described in  in Appendix \ref{sec:appendix_quality},
we follow up with GPT-4 quality pilot by annotating ten conversations with GPT-4 and asking the experts for the 5-point Likert scale evaluation (one overall score for ten conversations). The results are presented in Table \ref{tab:pilot_validation_GPT}.

 Some experts pointed out that sometimes the language seems ``stuffy'' and ``medical'', thus leading us to prompt refinement \footnote{We refined the prompt additional language consideration: \textit{Use professional and friendly language when giving feedback. Focus on what is most beneficial to hear.}. 
} The final prompt can be found in Appendix \ref{sec:appendix_prompt}.

\begin{table}[ht]

\centering
  \begin{tabular}{|c|c|}
  
    \hline
    \multirow{2}{*}{Evaluator} & \multicolumn{1}{c|}{Annotator} \\ \cline{2-2}
     & GPT-4\\ \hline
    Expert A &  5/5\\ \hline
    Expert B &  5/5\\ \hline
    Expert C  & 5/5\\ \hline
  \end{tabular}
  \caption{Quality validation pilot results for GPT-4 generated annotations.}
  \label{tab:pilot_validation_GPT}
\end{table}

\section{Expert-only vs. GPT-4+expert annotations}
\label{sec:appendix_Expert_vs_GPT4+Expert}
All experts annotated a set of ten conversations. The sets were different so that later evaluations are not biased by comparison to oneself, i.e., ``\textit{this is not good because I did something else}''\footnote{Due to the subjective nature of this task, there is no single correct way of annotating.}. Additionally, the experts annotated another set of ten conversations, this time, refining GPT-4 feedback (refinement instructions in Appendix \ref{sec:appendix_refinement}).

Each expert then evaluated the quality of annotations made by other experts with and without GPT-4 default feedback (7 questions asking about feedback components, 5-point Likert scale) and compared on utterance level whether expert-only annotation or GPT-4+expert annotation is preferred (or there is no significant difference) (assessment instructions in Appendix \ref{sec:appendix_comparison}).

\subsection{Do experts consistently validate themselves?}
\label{sec:appendix_final_validation_annotators}

\begin{figure}[th]
\centering
\includegraphics[width=3in]{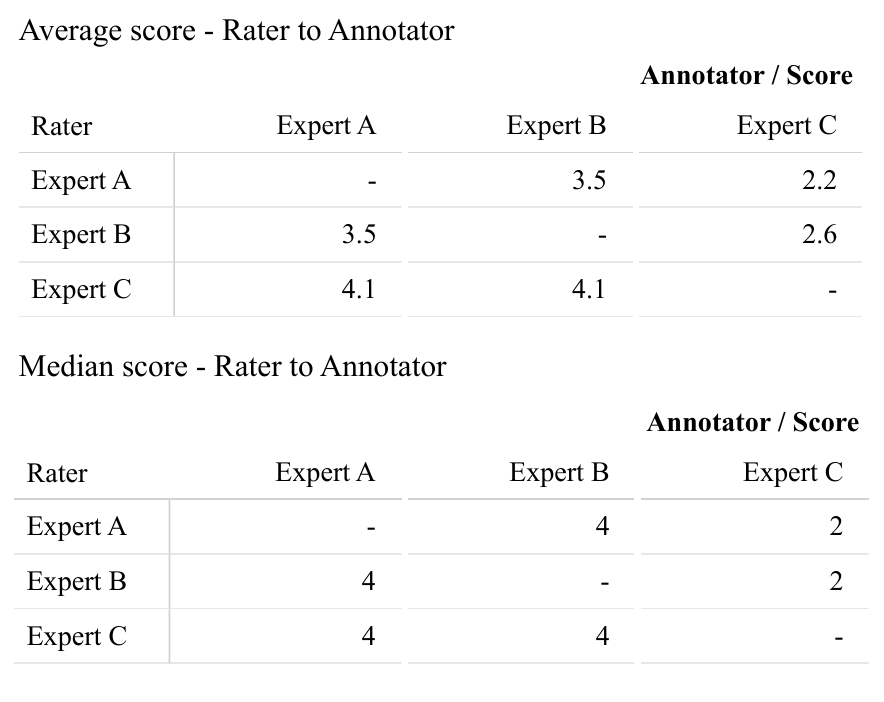}
\caption{Tables presenting average and median score for quality of annotations for experts A, B and C. Each entry in the table shows how a particular expert (row) rated another expert's annotation (column). The summary of each column provides the overall quality score of the expert's annotations.}
\label{fig:experts_validation}
\end{figure}

\begin{figure}[th]
\centering
\includegraphics[width=3in]{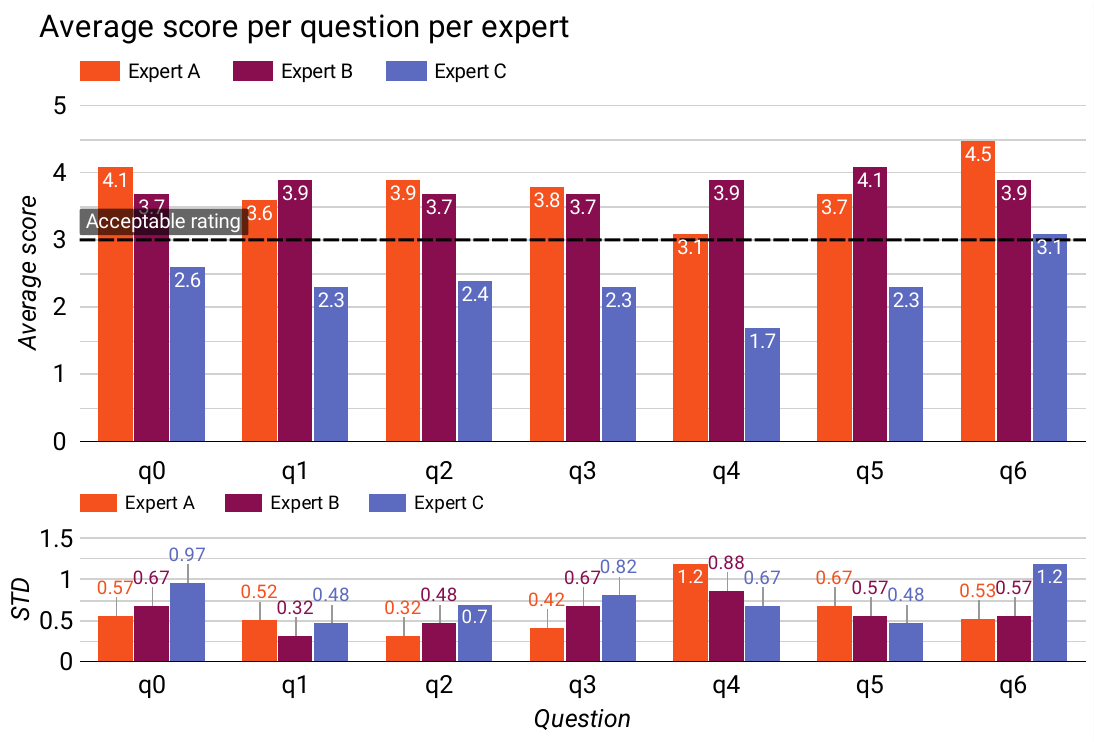}
\caption{Figure presents average score with standard deviation for each expert A, B, C broken down by 7 questions used to assess the quality of experts annotations. The dotted line marks the \textit{Acceptable} rating (3). }
\label{fig:experts_validation_per_question}
\end{figure}

Experts A and B get high ratings, even without GPT-4 pre-annotation. While expert C initially demonstrated the ability to produce high-quality annotations, there appears to be some inconsistency in maintaining the same level of quality across an entire batch of conversations (scores below \textit{Acceptable} rating). Figure \ref{fig:experts_validation} presents the average and median score of each expert rated by every other expert. 
Figure \ref{fig:experts_validation_per_question} presents how each expert overall (averaged over the raters) was scored in each of the questions asking about different feedback components -- the list of questions is attached at the end of the Appendix \ref{sec:appendix_comparison} presenting the interface used to conduct the study. 

Experts A and B consistently achieve scores around 4 which translates to \textit{Good} quality. Experts C fails to exceed the \textit{Acceptable} rating for all questions across the board. Moreover, their answers are also subject to the highest variation in the score in majority of cases. 

\subsection{Do annotations benefit from GPT-4 usage?}
\label{sec:appendix_benefit_GPT4}

With Expert C excluded as a rater, we compare the average annotations quality score of Expert A and B with and without GPT-4 pre-annotations (same setting as in the validation pilot - 7 questions and 5-point Likert scale).

The average score assessing the annotations' quality improves when GPT-4 is used for pre-annotations (Table \ref{tab:annotation_quality_gpt4_pilot}). Moreover, the standard deviation of the scores decreases. Taking these factors combined, the results point to higher and more consistent quality of annotations when GPT-4 is used. 

\begin{table}[h!]
\centering
    \begin{tabular}{l|c}
      \textbf{Annotation method} & \textbf{Average score} \\
      \hline
      Expert-only & \( 3.54 \pm 0.81 \) \\
      GPT-4 + Expert & \( 3.96 \pm 0.62 \) \\
    \end{tabular}
    \caption{Comparison of the average score of annotations' quality averaged over the experts without and with GPT-4 pre-annotations.}
\label{tab:annotation_quality_gpt4_pilot}
\end{table}

\begin{figure}[th]
\centering
\includegraphics[width=3in]{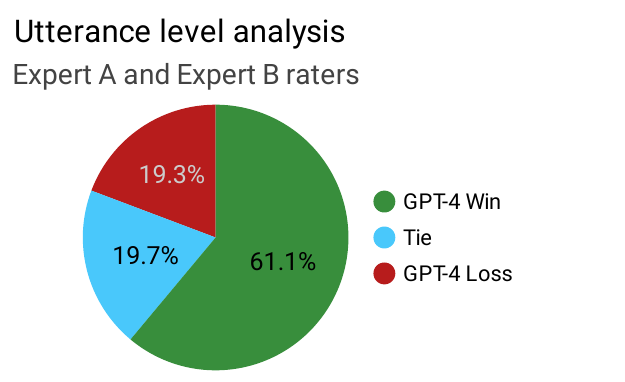}
\caption{Diagram presenting the distribution of whether the raters (Experts A and B) prefer annotations with or without GPT-4 pre-annotations. The figure presents percentages for three options: GPT-4 Win (green) -- 61.1\%, Tie (blue) -- 19.7\%, and GPT-4 Loss (red) -- 19.3\%.}
\label{fig:utt_lev_analysis}
\end{figure}

Additionally, GPT-4 + Expert is strongly preferred on the utterance level (see Figure \ref{fig:utt_lev_analysis}). Presented with two annotations, one with and the other without GPT-4 pre-annotations, raters in the majority of cases (61.1 \%) prefer the ones with pre-annotations. In 19.7\% of cases, they are indifferent, and in 19.3\% of cases, they prefer annotations without GPT-4 pre-annotations. 

Qualitatively, when experts refine annotations, they tend to add extra feedback components, for instance, adding an extra goal over the one already pointed out by GPT-4. They sometimes rephrase goal/alternative response chunks that can be improved (e.g., making the question more open-ended). Those were thus not only due to fixing errors but also aim to refine and follow individual preferences \cite{sripada2005evaluation}. 

We hypothesize that GPT-4 + Expert are preferred since they allow experts to focus on what is most important and refining parts where GPT-4 failed. This reduces the work burden of writing everything from scratch. Quantitatively, we conduct the Wilcoxon test \cite{wilcoxon1992individual} on conversation ratings, and statistically, GPT-4+Expert conversations obtain better ratings (p<0.05). The win/loss rate is also statistically significant (Wilcoxon and Binomial test, p <0.05). Sample examples of GPT-4 + Expert vs. Expert-only annotations are provided in Appendix \ref{sec:appendix_preference}.

Based on the above pilot results, we continue annotating QESConv with Experts A and B. 

\section{Fine-tuning experimental setup}
\label{sec:appendix_experimental_setup}

To curate a fine-tuning dataset, we leverage our FeedbackESConv data. To format each training datapoint we follow Alpaca style instruction formatting \cite{alpaca}. Each datapoint contains as output the feedback annotations from FeedbackESConv for the utterance $U_{i}$, with goal \& alignment parts preceding the alternative answer, to provide ``explanations'' in order to guide the generation process \cite{wei2022chain}.

The input is the conversation context $\mathrm{c}_{i}$. To find the part of the conversation to provide relevant context, we follow \cite{chen2020multi} by segmenting the conversation using C99 algorithm\cite{choi2000advances} on utterance embeddings. We embed the utterances using HuggingFace  \cite{huggingface} transformer model \textit{all-MiniLM-L6-v2}. We define the relevant context for each utterance as all past utterances in the current and previous segments. 

We fine-tune and align for 3 epochs. We use a single A100 GPU for the experiments. Overall, our computational budget amounted to approximately 130 GPU hours.

\newpage
\onecolumn
\includepdf[scale=0.8,pages=1,pagecommand={\section{Data annotation guide} \label{sec:appendix_annotation} \thispagestyle{empty}}, fitpaper=true]{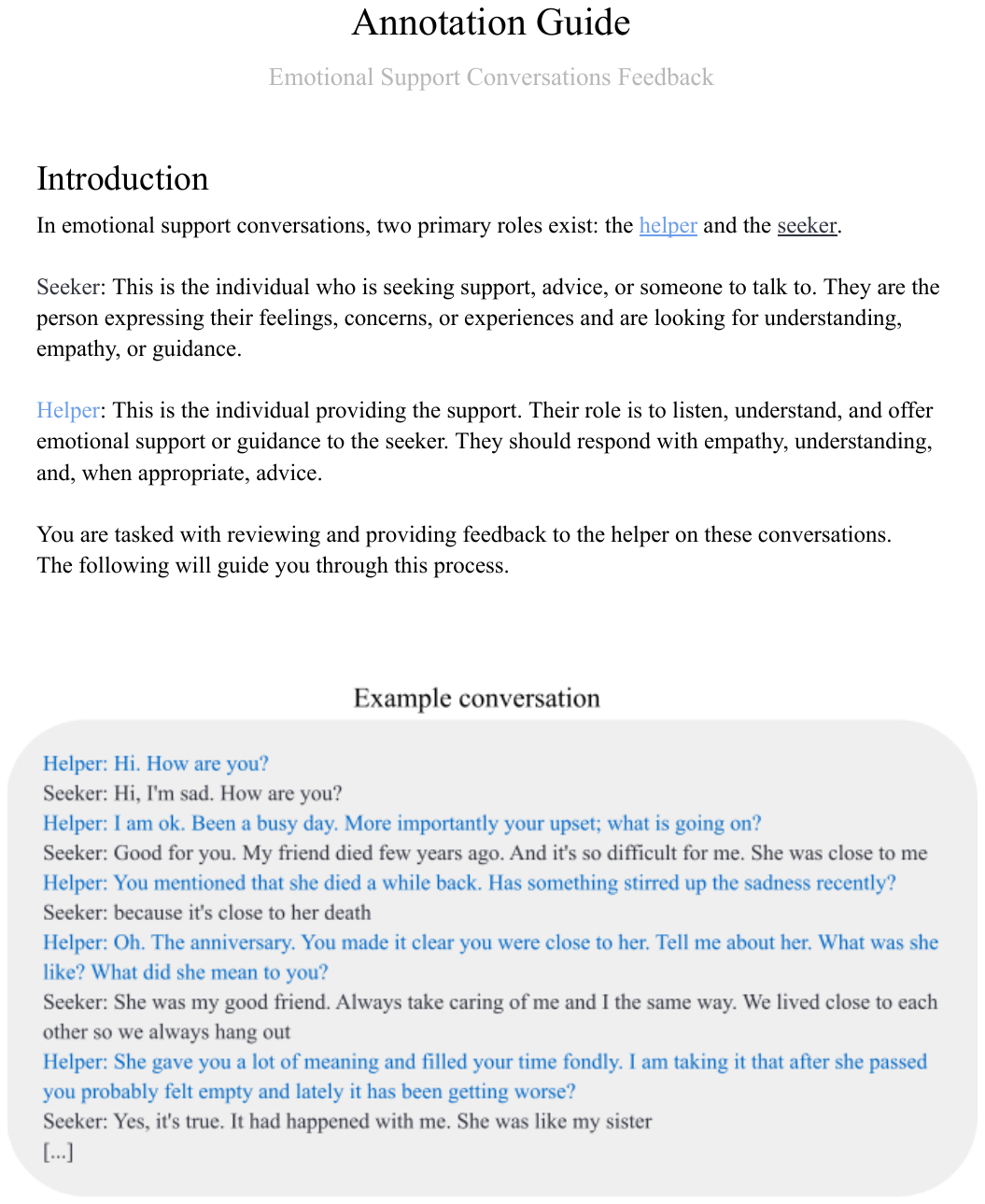}
\includepdf[scale=0.8,pages=2-,pagecommand={\thispagestyle{empty}}, fitpaper=true]{figures/guide.pdf}

\newpage
\onecolumn
\includepdf[scale=0.8,pages=1,pagecommand={\section{GPT4 Prompt \& In-context Learning} \label{sec:appendix_prompt} \thispagestyle{empty}}, fitpaper=true]{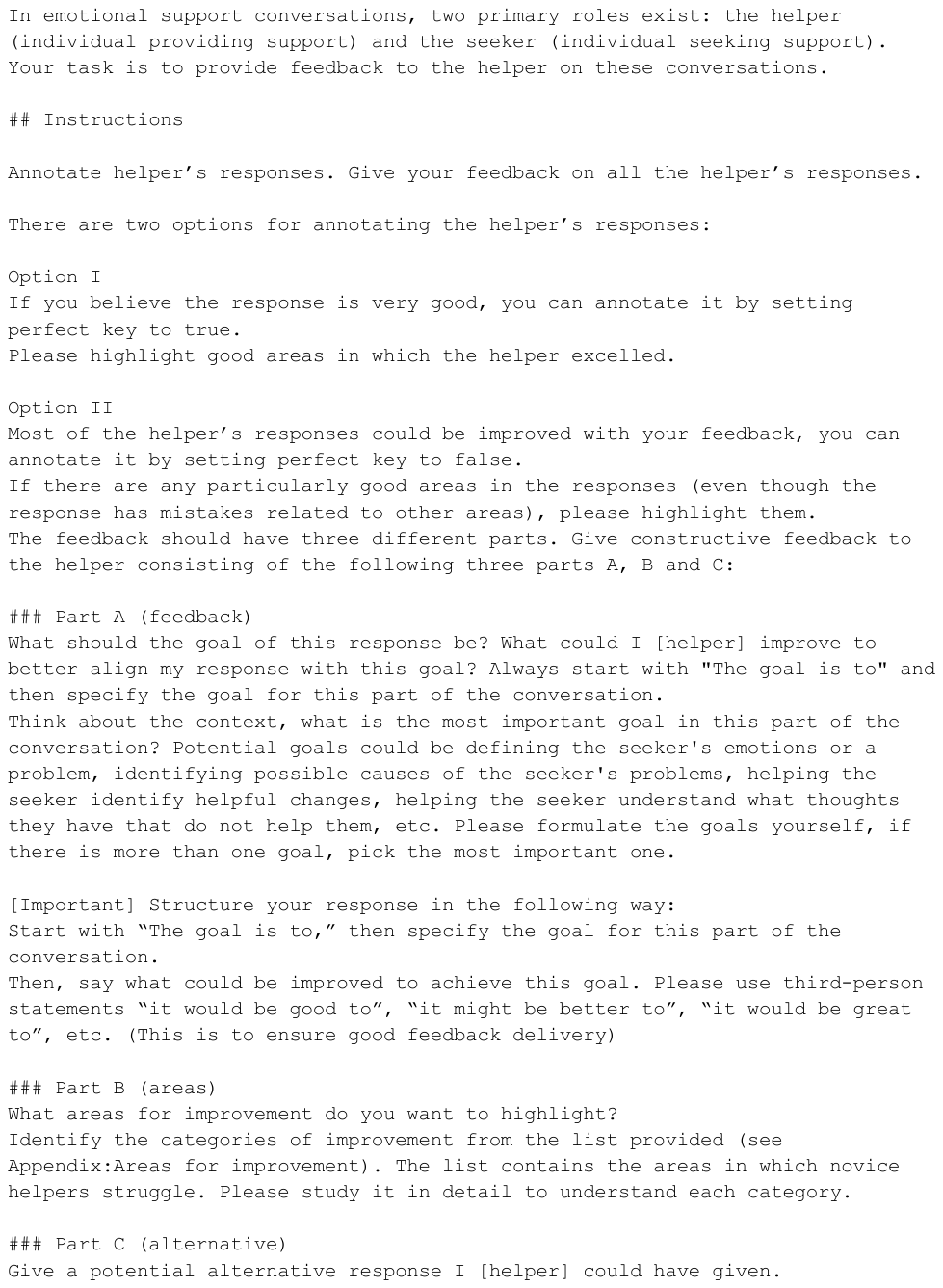}
\includepdf[scale=0.8,pages=2-,pagecommand={\thispagestyle{empty}}, fitpaper=true]{figures/GPT4_prompt.pdf}

\newpage
\onecolumn
\includepdf[scale=0.8,pages=1,pagecommand={\section{Data annotation with refinement guide} \label{sec:appendix_refinement} \thispagestyle{empty}}, fitpaper=true]{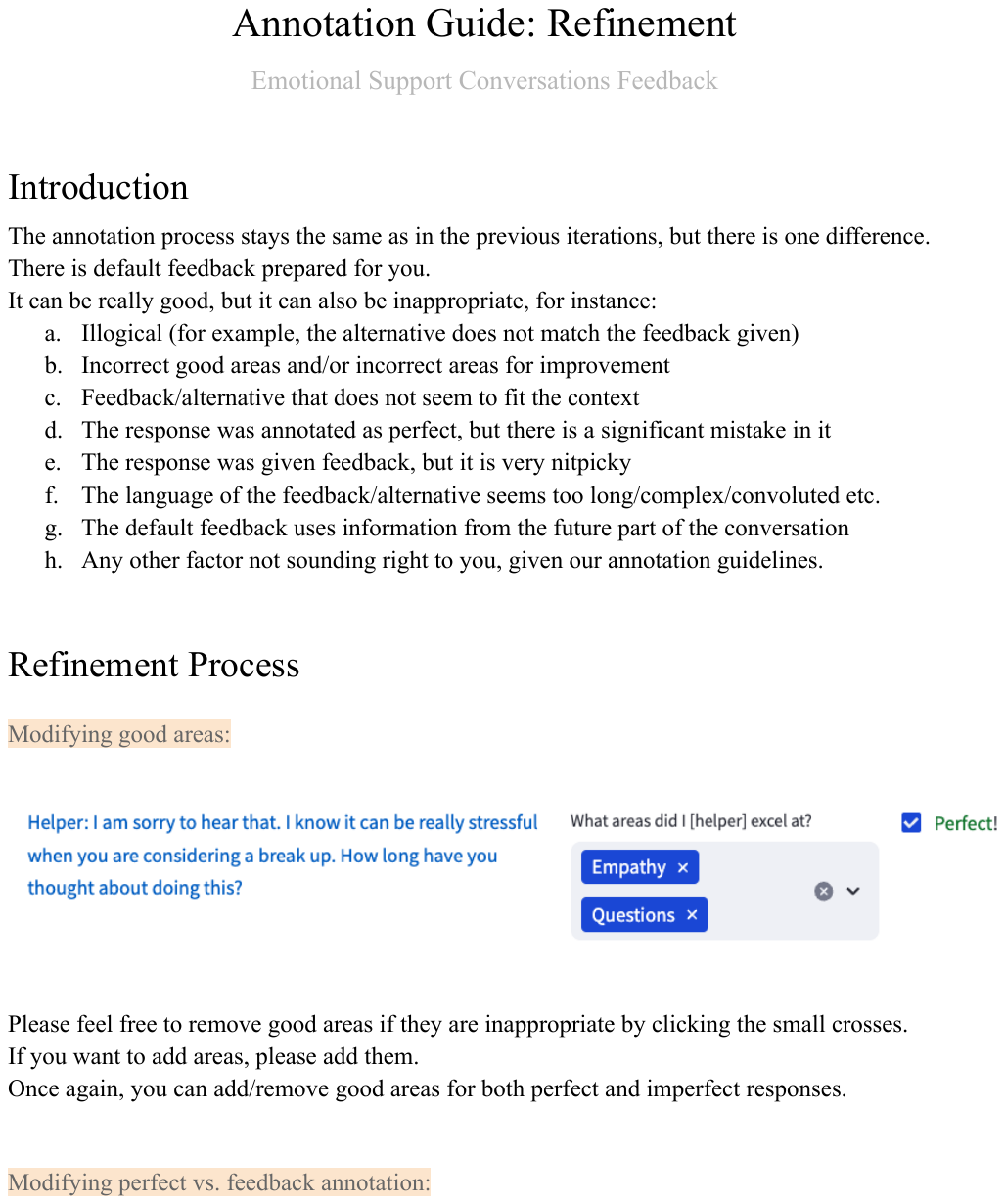}
\includepdf[scale=0.8,pages=2-,pagecommand={\thispagestyle{empty}}, fitpaper=true]{figures/guide_refinement.pdf}

\newpage
\onecolumn
\includepdf[scale=0.8,pages=1,pagecommand={\section{Annotation comparison} \label{sec:appendix_comparison} \thispagestyle{empty}}, fitpaper=true]{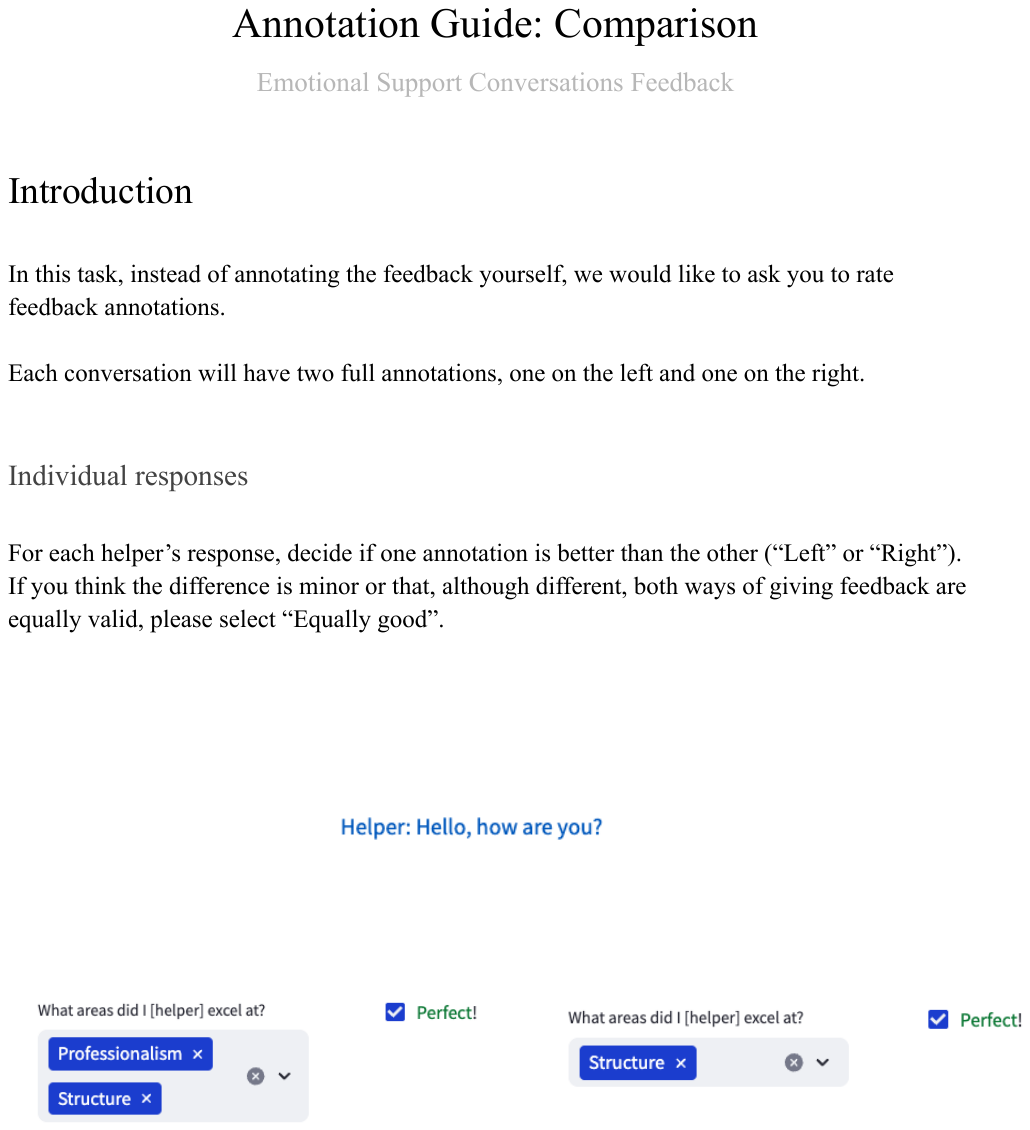}
\includepdf[scale=0.8,pages=2-,pagecommand={\thispagestyle{empty}}, fitpaper=true]{figures/guide_comparison.pdf}

\newpage
\onecolumn
\includepdf[scale=0.9,pages=1,pagecommand={\section{GPT-4 vs. Expert only examples} \label{sec:appendix_preference} \thispagestyle{empty}}, fitpaper=true]{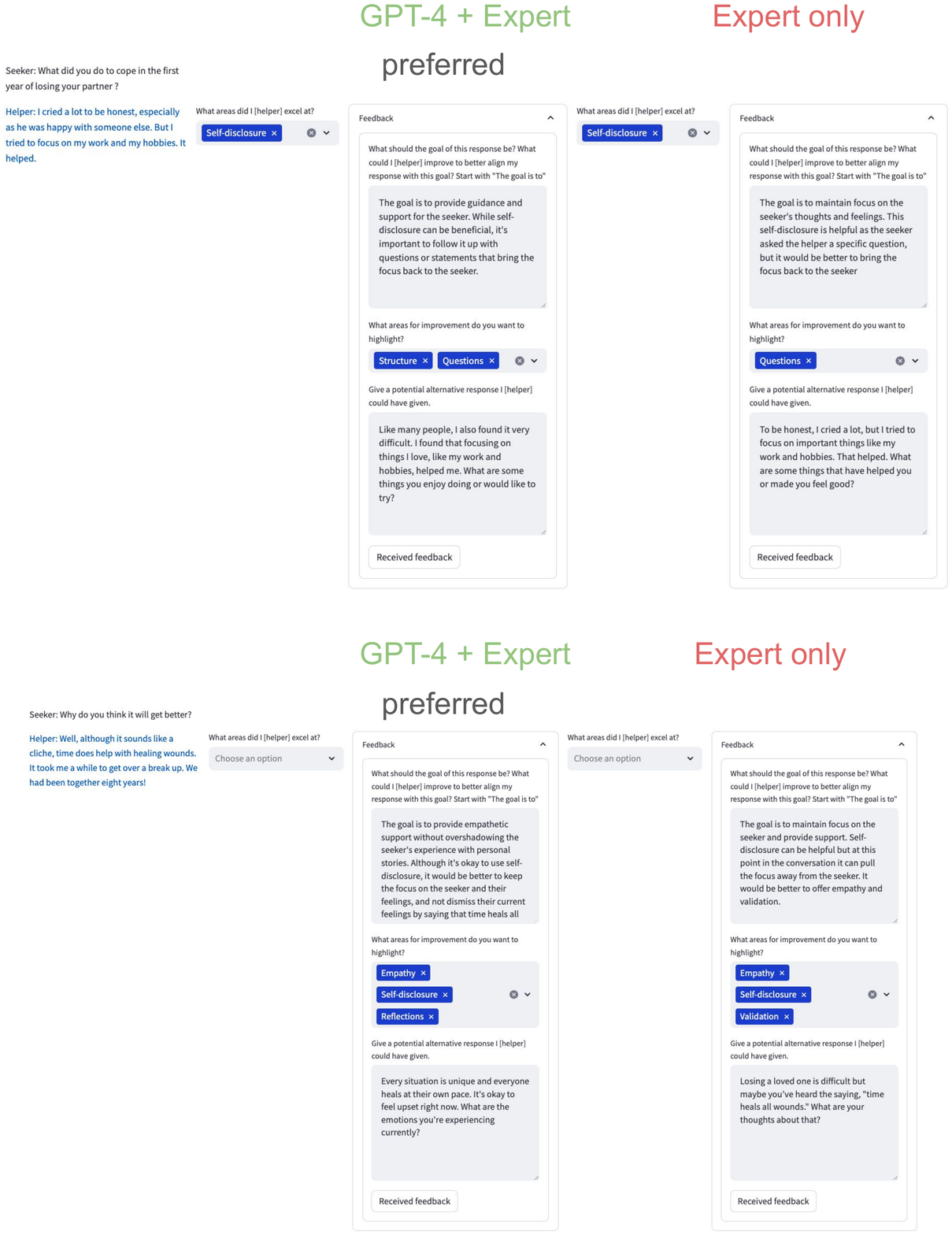}
\includepdf[scale=0.9,pages=2-,pagecommand={\thispagestyle{empty}}, fitpaper=true]{figures/preference.pdf}

\newpage
\onecolumn
\includepdf[scale=0.9,pages=1,pagecommand={\section{Feedback samples generated by the $\mathcal{M}_{\text{Self-imp}}$ model} \label{sec:appendix_generations} \thispagestyle{empty}}, fitpaper=true]{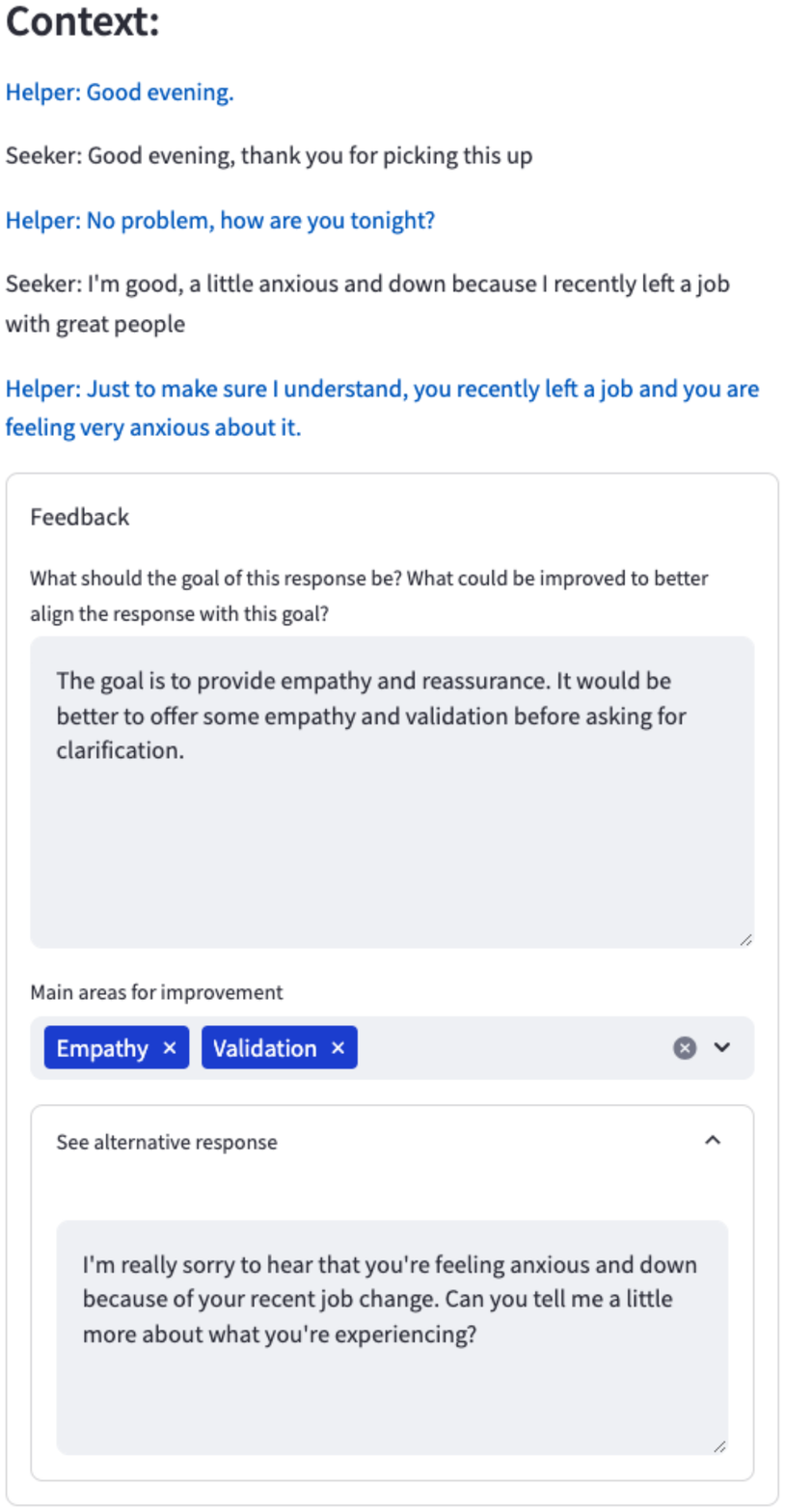}
\includepdf[scale=0.9,pages=2-,pagecommand={\thispagestyle{empty}}, fitpaper=true]{figures/example_generations.pdf}

\end{document}